\documentclass[10pt,twocolumn,letterpaper]{article}

\usepackage{cvpr}      

\definecolor{cvprblue}{rgb}{0.21,0.49,0.74}
\usepackage[pagebackref,breaklinks,colorlinks,allcolors=cvprblue]{hyperref}

\def\paperID{*****} 
\def\confName{CVPR}
\def\confYear{2026}

\usepackage[utf8]{inputenc}

\usepackage{amssymb}
\usepackage{amsmath}
\usepackage{url}
\usepackage{booktabs}
\usepackage{svg}

\usepackage{tabularx}
\usepackage{lipsum}
\usepackage{subcaption}
\usepackage{multirow}
\usepackage{booktabs}
\usepackage{adjustbox}
\usepackage{placeins} 
\usepackage{stfloats}
\newcolumntype{C}{>{\centering\arraybackslash}X}

\usepackage{hyperref}

\usepackage[table,xcdraw]{xcolor}

\usepackage{colortbl}
\usepackage{graphicx}

\usepackage[percent]{overpic}
\usepackage{tikz}  
\usepackage{pgfplots}
\pgfplotsset{compat=1.18}

\title{Inhibited Self-Attention:\\Sharpening Focus in Vision Transformers} 
\author{Peter R.D. van der Wal$^{1}$\thanks{Corresponding author: p.r.d.van.der.wal@rug.nl} \quad Nicola Strisciuglio$^{2}$ \quad George Azzopardi$^{1,3,4}$\\
$^1$University of Groningen \quad $^2$University of Twente \quad $^3$Stellenbosch University \quad $^4$University of Malta
}

\begin{document}
\maketitle

\begin{abstract}
Vision Transformers (ViTs) have demonstrated remarkable performance in computer vision tasks. However, their self-attention mechanism often diffuses focus across background regions, relying on spurious correlations rather than object-relevant cues. Inspired by inhibitory mechanisms observed in biological vision systems, we propose the Inhibited Self-Attention (ISA), a novel self-attention that integrates inhibitory signals to enhance feature selectivity and suppress spurious responses. In contrast to conventional self-attention, which relies solely on positive attention values due to softmax normalization, our approach retains and utilizes negative attention scores to suppress irrelevant features and sharpen focus on objects of interest. Experiments across multiple datasets, including ImageNet-1k and COCO, and several robustness benchmarks demonstrate that ISA enhances object-centric selectivity, reduces shortcut reliance, and improves out-of-distribution generalization. Our analysis of relevance maps confirms that ViTs with ISA exhibit sharper, more localized focus on object-relevant regions while reducing distractions from non-relevant (background) features, enabling more reliable models. We release our code at \url{https://github.com/prdvanderwal/inhibited-self-attention}.
\end{abstract}




\vspace{-0.2cm}
\section{Introduction}
Vision Transformers (ViTs) \cite{dosovitskiy_image_2021} have sparked a paradigm shift in computer vision, challenging the decade-long dominance of Convolutional Neural Networks (CNNs) ~\cite{krizhevsky_imagenet_2012, szegedy2015going, he2016deep}. Unlike CNNs, which rely on local feature extraction through convolutional filters, ViTs process an image as a set of patches, using self-attention to model global dependencies ~\cite{dosovitskiy_image_2021, li_localvit_2021, mao_towards_2022}. This enables ViT architectures to drop the inductive bias of convolutional operators, and use self-attention to expand the range of the receptive field to long-range relations among image patches. Remarkably, ViTs achieved higher performance than CNNs when trained on very large datasets, leading to foundation models that can be adapted to various tasks~\cite{dosovitskiy_image_2021}.

\begin{figure}[t]
    \centering
    \begin{subfigure}{0.425\textwidth}
        \includegraphics[width=\textwidth]{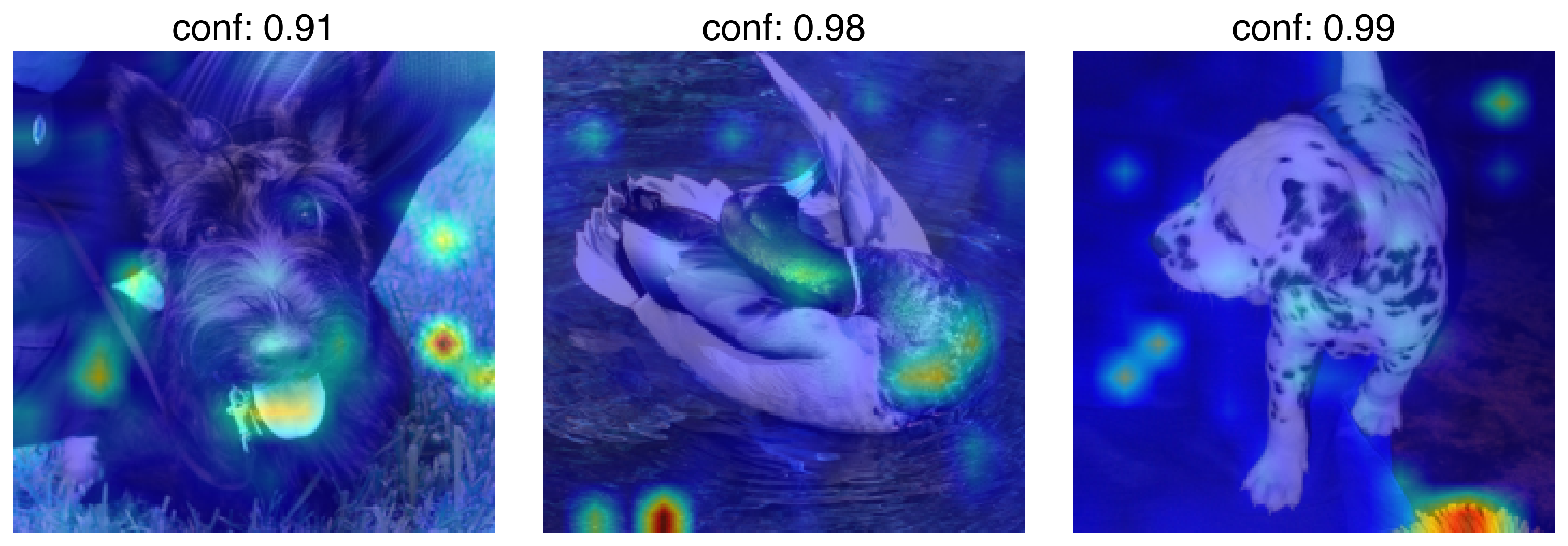}
        \subcaption{ViT-S/16} \label{fig:introA}
    \end{subfigure}
    \begin{subfigure}{0.425\textwidth}
        \includegraphics[width=\textwidth]{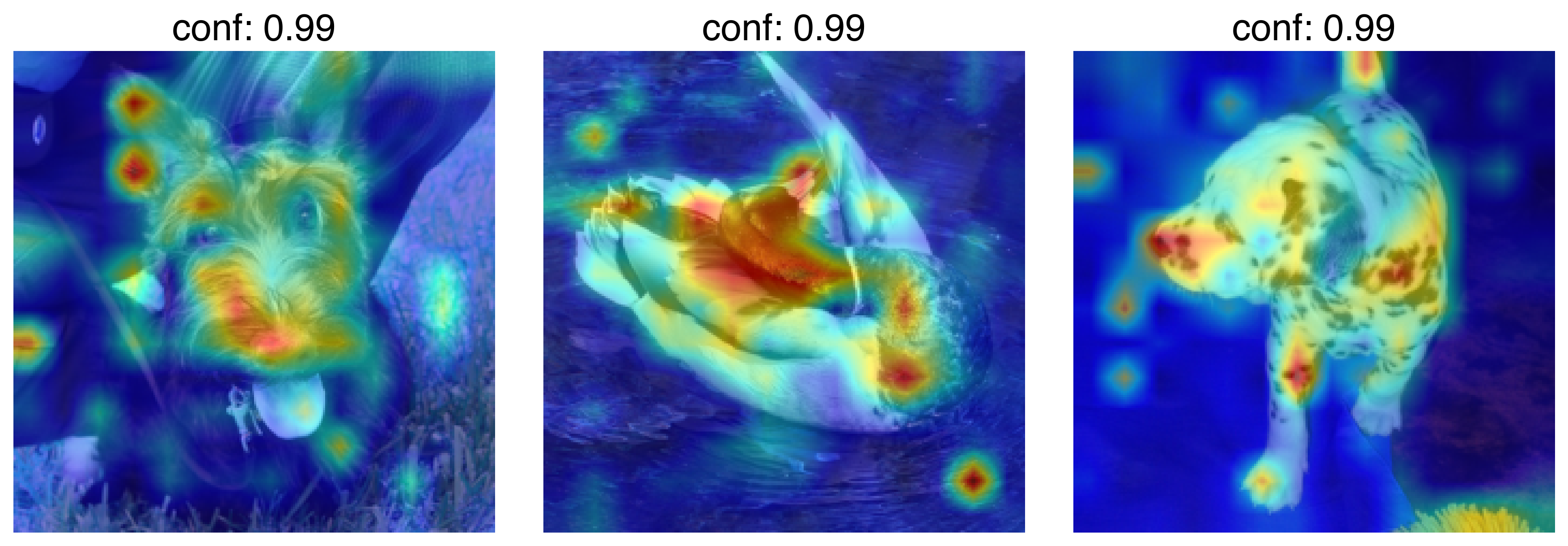}
        \subcaption{ISA-ViT-S/16} \label{fig:introNC}
    \end{subfigure}

    \caption{(a) Examples of correctly classified images by a ViT-S/16 with very high confidence where the attention of the model is not directed at the object of interest. (b) The same images classified by ISA-ViT-S/16 using our proposed inhibited self-attention.}
    \label{fig:intro}
    \vspace{-0.3cm}
\end{figure}

Despite their strong performance, ViTs do not always concentrate attention on the most informative image regions. They frequently distribute attention across background elements that contain easily detected but non‑informative spurious cues. This can introduce bias by over-relying on background features and reducing the ability of ViTs to learn class-specific representations \cite{9711024}. This limitation increases the likelihood of relying on incidental correlations rather than meaningful object attributes \cite{grisi2024maskedattentionmechanismimproving}.  In~\cref{fig:intro}, we show images correctly classified by a ViT with high confidence but with attention focused on the background or other shortcuts rather than on the objects of interest. Steering attention towards objects of interest, has been an active line of research in CNN based models, but comparatively fewer works have addressed this challenge in the context of ViTs. %

In this paper, we reshape the self-attention mechanism of Vision Transformers by incorporating an inhibitory process that enhances the selectivity of the model attention. Our method draws inspiration from biological visual processing, where inhibitory receptive fields in the visual system suppress responses to spurious stimuli, thus reinforcing the response to the patterns of interest~\cite{hubel_receptive_1962, nelson1978orientation, blakemore1972lateral}. We refer to this approach as \textit{Inhibited Self-Attention} (ISA). Unlike prior methods that rely on auxiliary models~\cite{ma2023rectify,tokselect} or add extra architectural branches (e.g., the Differential Transformer~\cite{ye2024differentialtransformer}), we make use of the already available negative attention scores, normally discarded by the softmax in standard Self-Attention, to introduce contextual inhibition and refine the final attention distribution. This mitigates a key limitation of conventional ViTs, which often allocate attention to spurious background features while reducing focus on class-relevant regions. By suppressing the non-relevant feature information, our approach enhances the ability of ViT models to learn more discriminative and meaningful feature representations. We investigate the improvement of the selectivity and robustness of ViTs brought by our method in object classification and detection, and particularly in the presence of shortcuts and 
distribution shifts.

Our \textbf{main contributions} are: (1) We introduce a novel inhibited self-attention mechanism that incorporates negative attention values into ViTs, which contribute to sharpening the focus of self-attention on relevant image regions, without the overhead or architectural complexity of auxiliary models. (2) We introduce a new metric called `Attention-on-Objects' to quantify object-centered attention in ViTs. 

\section{Related Work}

\paragraph{ViTs and Softmax in Self-Attention.}
Inspired by the transformer model for natural language processing applications~\cite{vaswani_attention_2017}, the Vision Transformer (ViT)~\cite{dosovitskiy_image_2021} has caused a paradigm shift in the field of computer vision. In contrast to CNNs which work on the locality principle with filters that only observe a region of the input image, the receptive field of ViTs is effectively expanded to the entire image through the self-attention mechanism~\cite{li_localvit_2021}. This allows ViTs to capture rich and highly flexible global patterns~\cite{tokselect}. In self-attention, each image patch is mapped to three vectors: queries, keys, and values. Similarity scores between queries and keys are computed, which can be positive or negative depending on the relationship between the patches. These raw scores are then scaled and passed through the softmax activation function, which normalizes them into a probability distribution. The softmax function converts both positive and negative similarities into positive weights, thereby eliminating the inhibitory effect of negative correlations and limiting the model's ability to capture relational structures involving both excitatory and inhibitory interactions.

In an effort to combine the global connectivity of ViTs with the locality of CNNs, several variants have been proposed, including Swin Transformer~\cite{liu_swin_2021}, ScopeViT~\cite{nie2024scopevit}, LocalViT~\cite{li_localvit_2021}, and ReViT \cite{diko2024revit}. For object detection, DETR~\cite{carion2020end} introduced an end-to-end transformer framework later refined by DINO~\cite{zhangdino} with denoising queries and improved feature alignment. Data-efficient Image Transformers (DeiT)~\cite{touvron_training_2021} addressed the need for costly pre-training through distillation, using a teacher classifier and a dedicated distillation token that interacts with all embeddings. More recently, it was shown that competitive ImageNet-1k performance can also be achieved when training from scratch~\cite{beyer2022better}.

\paragraph{Biologically-Inspired Vision Models}
Biological neural networks have often served as inspiration for artificial neural networks (ANNs), particularly in vision research. Similar to a CNN that learns a filter to only respond to particular stimuli, the majority of the neurons in the V1 area of the visual system of the brain display a similar behaviour called orientation selectivity \cite{hubel_receptive_1962}. Some neurons in area V1 exhibit a response inhibition mechanism, called push-pull inhibition~\cite{hubel_receptive_1962, ferster_orientation_1996}, that sharpens the selectivity by suppressing responses to
 non-preferred stimuli, such as spurious texture. A computational model of simple cells for image processing was proposed in~\cite{azzopardi_corf_2012}, and subsequently extended by including the push-pull inhibition mechanism in~\cite{azzopardi_push-pull_2014}, with improved performance in noisy images. 
 Other key inhibition mechanisms include center-surround inhibition, which enhances edge detection by suppressing uniform backgrounds ~\cite{nelson1978orientation,grigorescu2003contour}, and lateral inhibition, which sharpens contrast by reducing redundant responses by neighboring neurons~\cite{blakemore1972lateral}.
Lateral inhibition was exploited in AlexNet~\cite{localresponseAlex}, using a Local Response Normalization, yielding improved generalization performance. Furthermore, in~\cite{strisciuglio_enhanced_2020,bennabhaktula2024pushpull}
push-pull inhibition was deployed in a new type of CNN layer, improving the robustness of models to image corruptions. 

The implementation of an inhibition mechanism in \cite{yao_bio-inspired_2023} that functions in parallel with components of the Swin-Transformer \cite{liu_swin_2021}, obtained improved accuracy, sensitivity, and specificity in comparison with the regular Swin-Transformer on a medical image dataset. 

In this work, we propose a novel self-attention mechanism that exploits negative attention scores to inhibit the attention of non-relevant features, thus sharpening the reliance on relevant areas of the objects of interest.

\paragraph{Enhancing Attention Selectivity}
Despite many variations of the self-attention mechanism, ViTs are prone to attend to irrelevant parts of the image such as background features~\cite{xue_vision_2022, ye2024differentialtransformer}, risking low selectivity of the models on the features that strictly characterize the patterns of interest. This phenomenon was empirically demonstrated in~\cite{tokselect} and identified as a leading reason for the training difficulties of ViTs. 

While object-centric mechanisms have been widely investigated for CNNs~\cite{li2018tell, wang2019sharpen, 9284467}, their direct adaptation to attention in Vision Transformers has received comparatively less focus.
In order to improve the selectivity of ViT models for relevant features of the objects of interest, in~\cite{tokselect} the authors proposed a convolution-based token-selector block that can be inserted right after the attention mechanism. In~\cite{xue_vision_2022} a CNN extractor was used to select relevant patches, such that only the attention over these patches is computed. This approach improves the selectivity and reduces the computational costs (as the self-attention is computed between a lower number of patches) of ViTs. However, its performance relies on the quality of the CNN-based patch descriptor. Similarly, in \cite{ma2023rectify} the authors aim to rectify shortcut learning in ViTs but rely on a separate saliency model. Instead of using an auxiliary model for patch selection or saliency calculation, in~\cite{mao_towards_2022} a position-aware attention scaling was introduced. 

Recent work has likewise identified that softmax discards information from negative query–key interactions. PolaFormer~\cite{meng2025polaformer} recovers these components within linear attention, where non-negative feature maps are required for a valid kernel decomposition, by routing same-signed and opposite-signed interactions through parallel streams, motivated by expressivity and efficiency rather than selectivity. Cog Attention~\cite{lv2024more} allows negative output weights by multiplying softmax applied to absolute QK scores with the sign of the raw QK products, framing negativity as a dynamic token-deletion mechanism targeting expressivity in language modelling and diffusion models, without addressing object-centric selectivity or robustness. The Differential Transformer~\cite{ye2024differentialtransformer} was proposed for language modeling to reduce attention noise through subtractive filtering between two softmax branches; this design requires two separate query–key projection pairs per layer, which increases parameter overhead, and is less suited to vision tasks with large token counts. Furthermore, unlike ISA, it operates primarily at the attention output level, which may be insufficient to produce the highly selective attention focus required to mitigate shortcut reliance. The Integral Transformer~\cite{kobyzev-etal-2025-integral} instead averages multiple QK logit signals before softmax, trading subtraction for integration, but reduces the per-signal head dimension by a factor equal to the number of signals, lowering the expressiveness of projections.

ISA utilizes a competitive interaction between two normalizations of the same attention signal (softmax and softmin), leveraging the negative attention information discarded by standard self-attention into an explicit inhibitory signal that sharpens attention and reduces noise. Unlike existing methods that require auxiliary models or extra query-key pairs, ISA is a seamless drop-in replacement for standard attention, with no additional parameter or FLOPs.

\section{Method: Inhibited Self-Attention}
\begin{figure*}[t]
\centering
\begin{minipage}{.2\textwidth}
\begin{subfigure}{\textwidth} 
\includegraphics[width=0.9\textwidth]{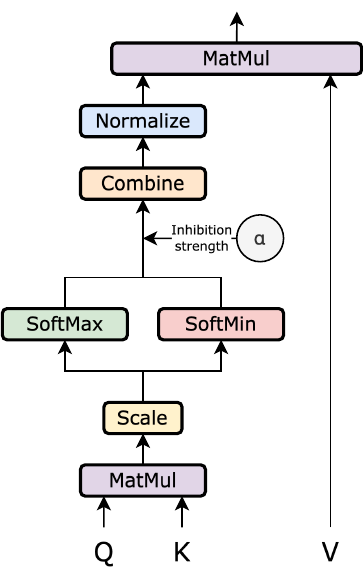}
\subcaption{Inhibited Self-Attention (ISA)}\label{fig:2a}
\end{subfigure}
\end{minipage}
\begin{minipage}{.70\textwidth}
\centering

\begin{subfigure}{\linewidth} 
        \centering
        \begin{tikzpicture}
            \node[font=\scriptsize] at (0, 3.25) {Image};
            \node[font=\scriptsize] at (2, 3.25) {Softmax};
            \node[font=\scriptsize] at (4, 3.25) {Softmin};
            \node[font=\scriptsize] at (6, 3.25) {ISA};

            \node at (3, 3.1) {\tikz \draw (0,0) -- (8,0);};

            \node at (0,2) {\includegraphics[width=2cm]{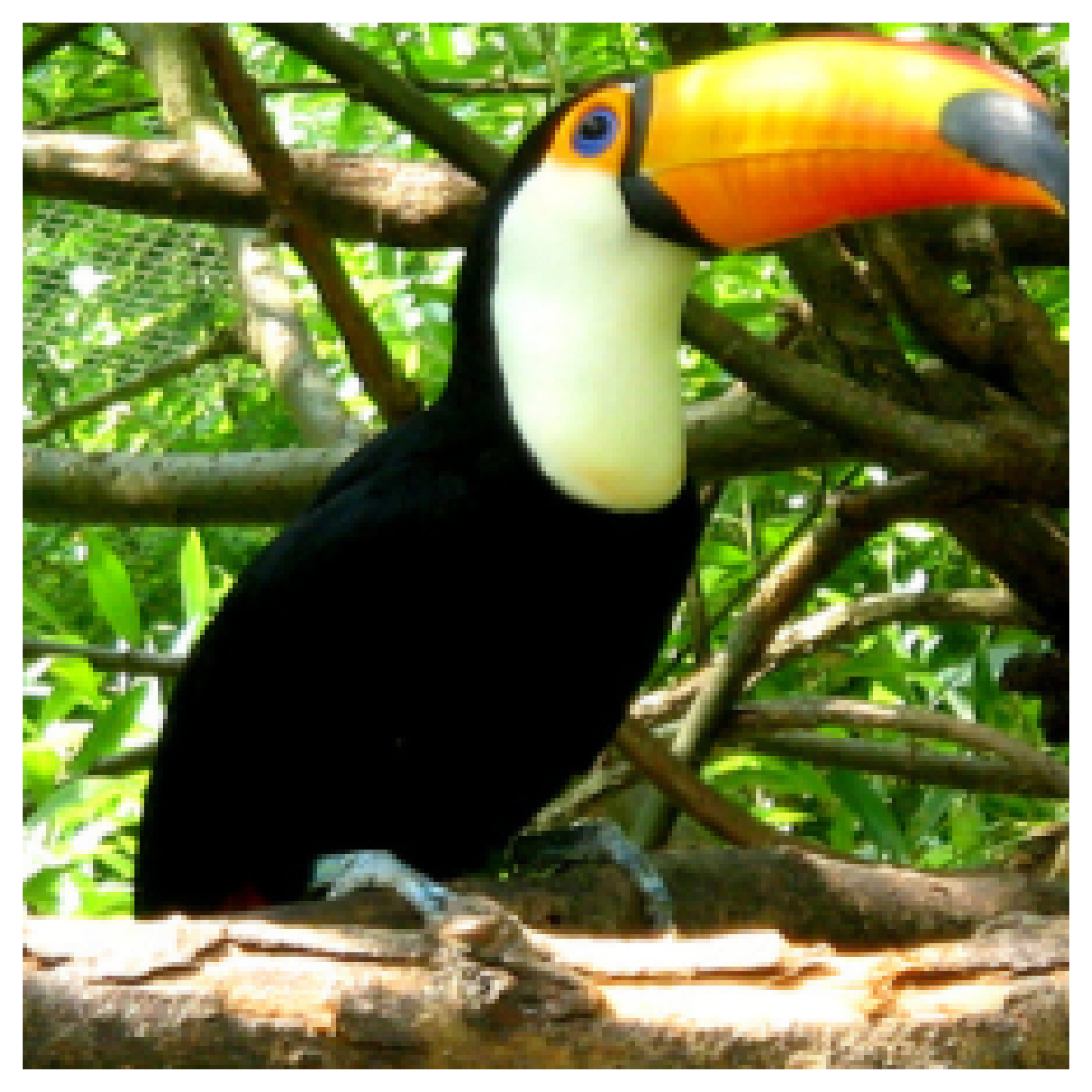}};
            \node at (2,2) {\includegraphics[width=2cm]{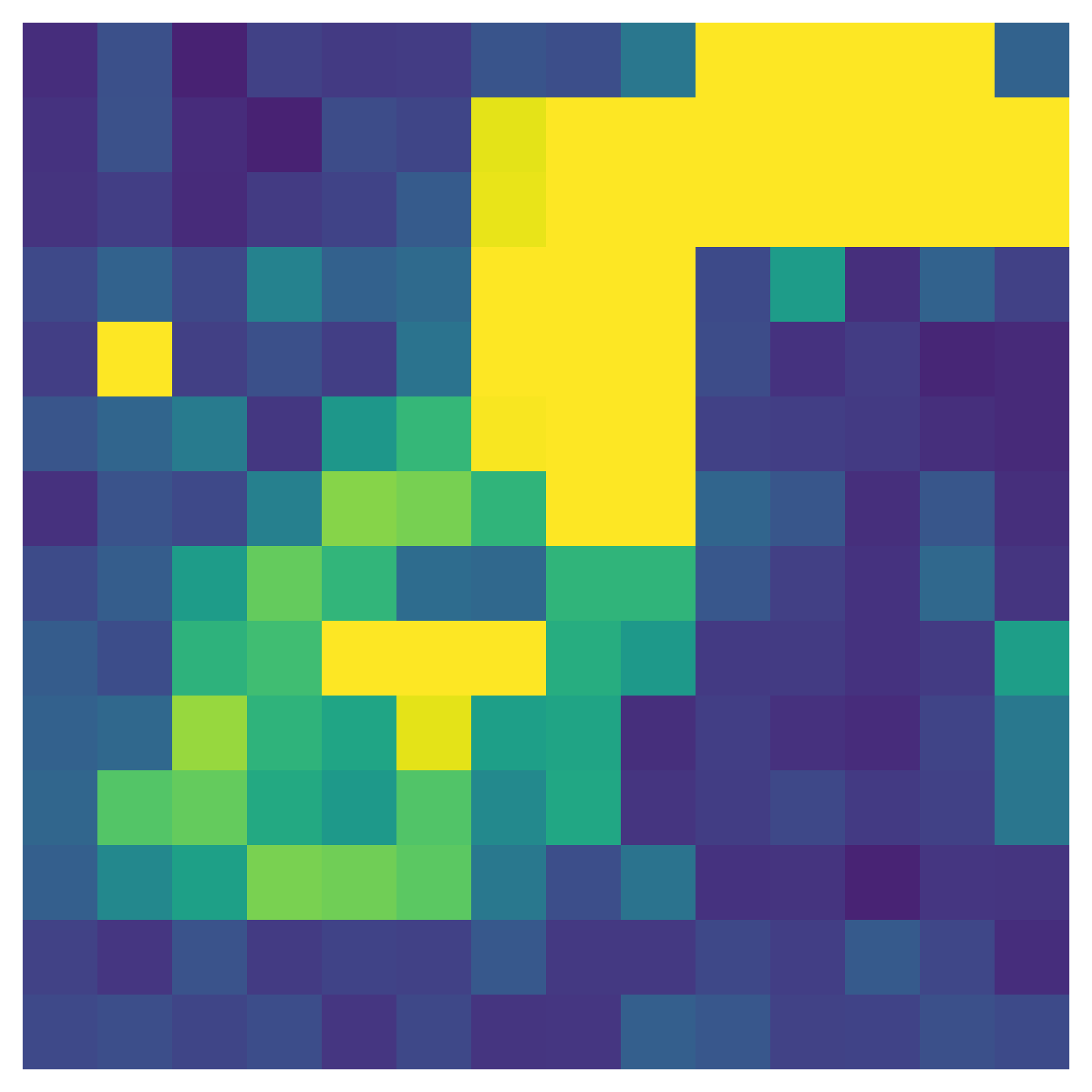}};
            \node at (4,2) {\includegraphics[width=2cm]{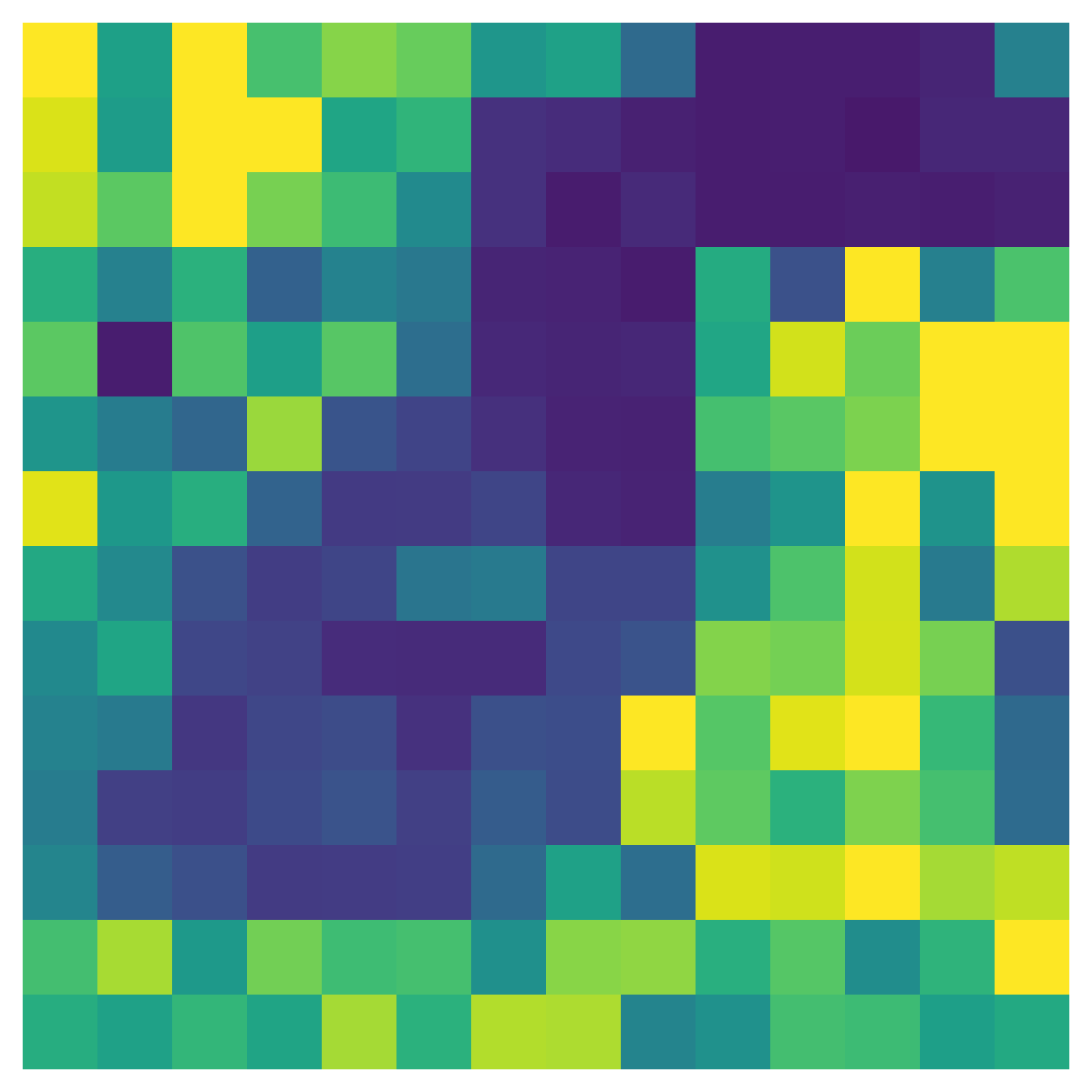}};
            \node at (6,2) {\includegraphics[width=2cm]{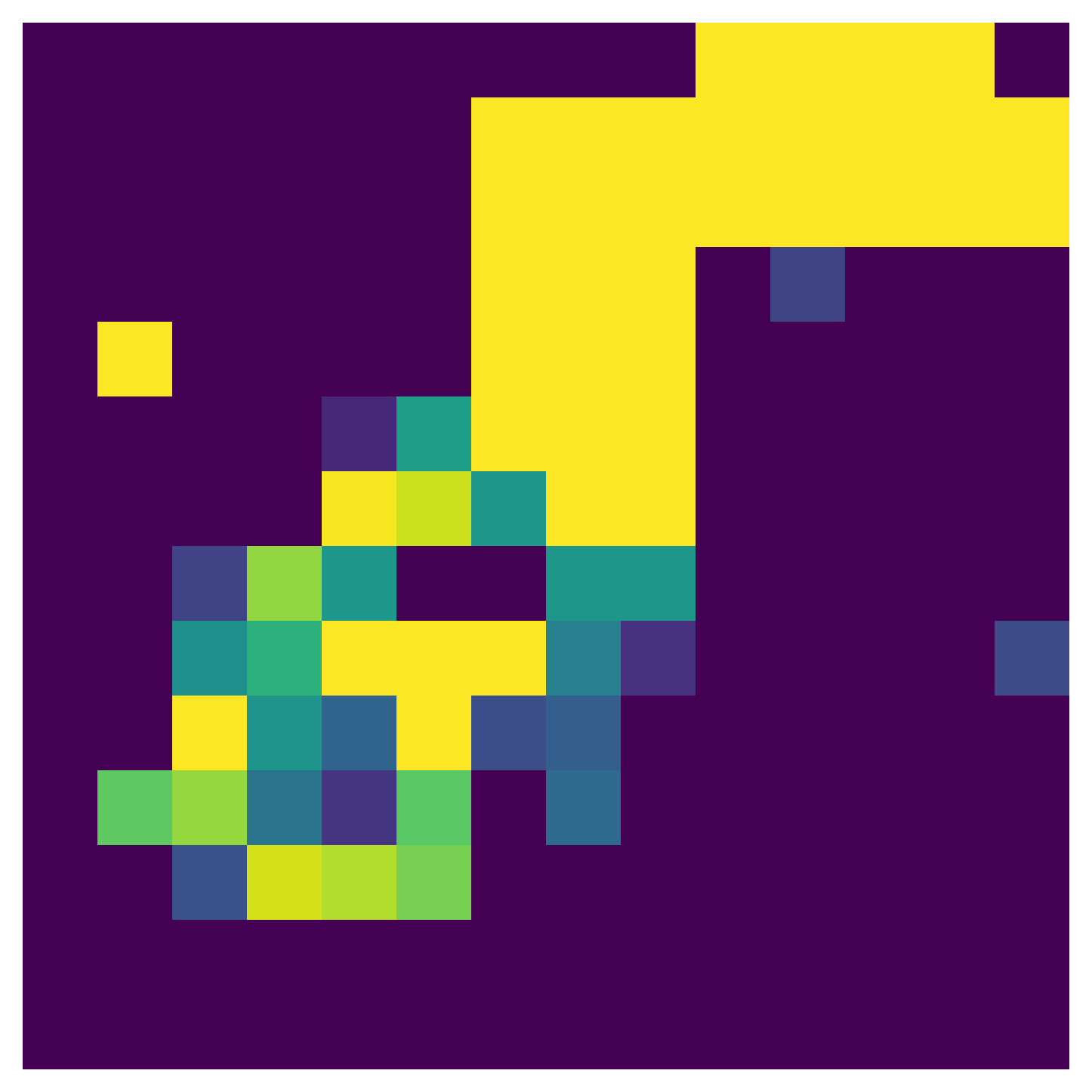}};

            \node at (0,0) {\includegraphics[width=2cm]{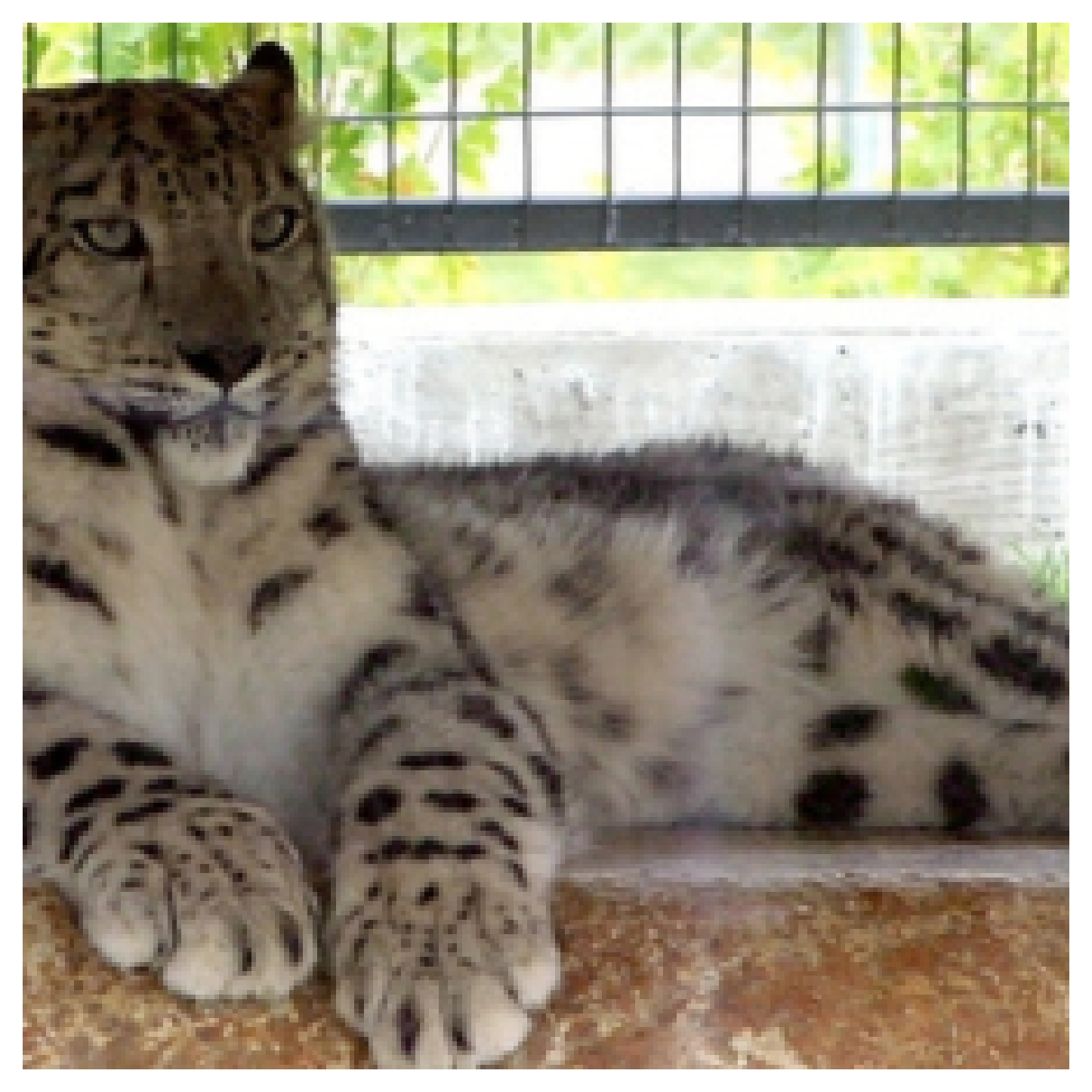}};
            \node at (2,0) {\includegraphics[width=2cm]{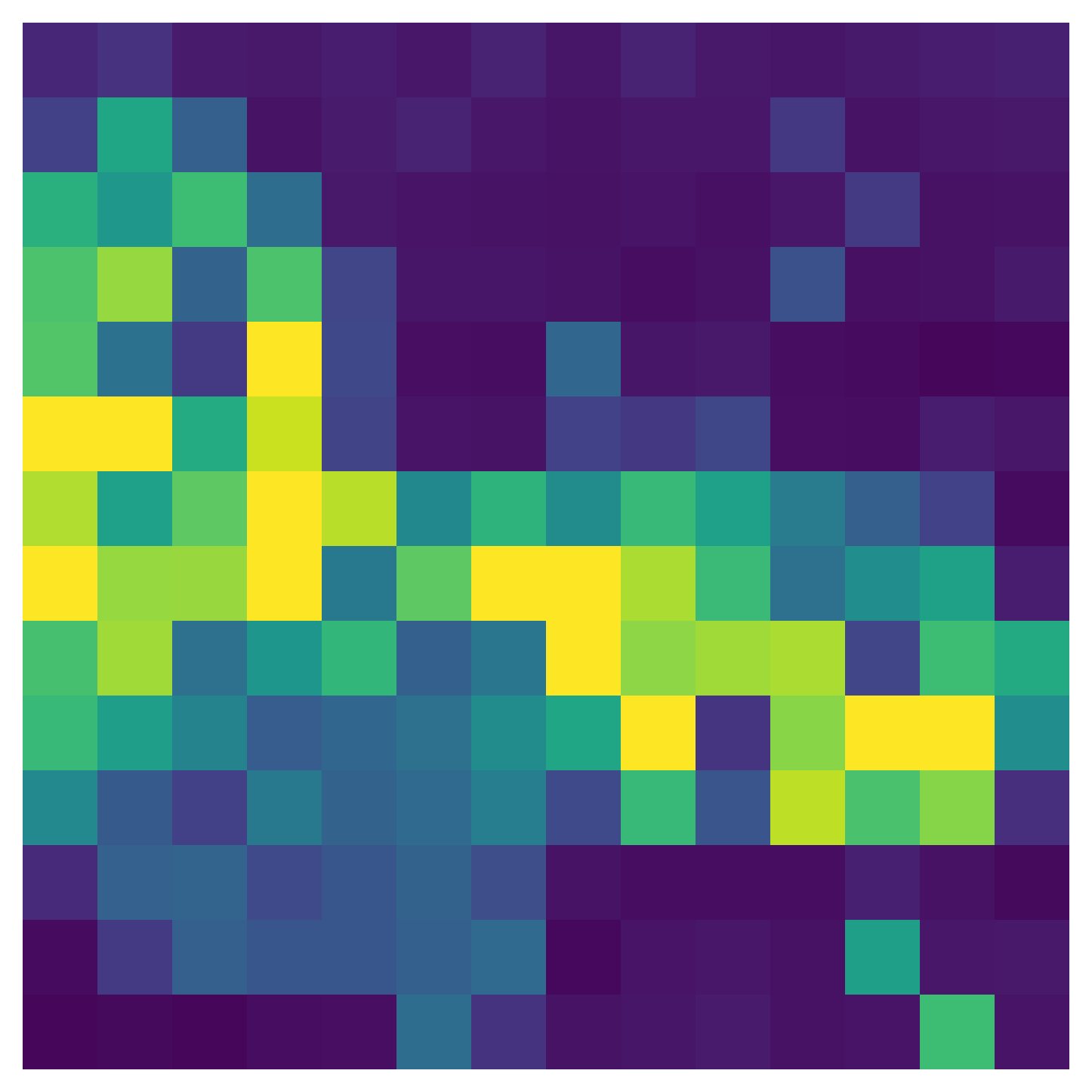}};
            \node at (4,0) {\includegraphics[width=2cm]{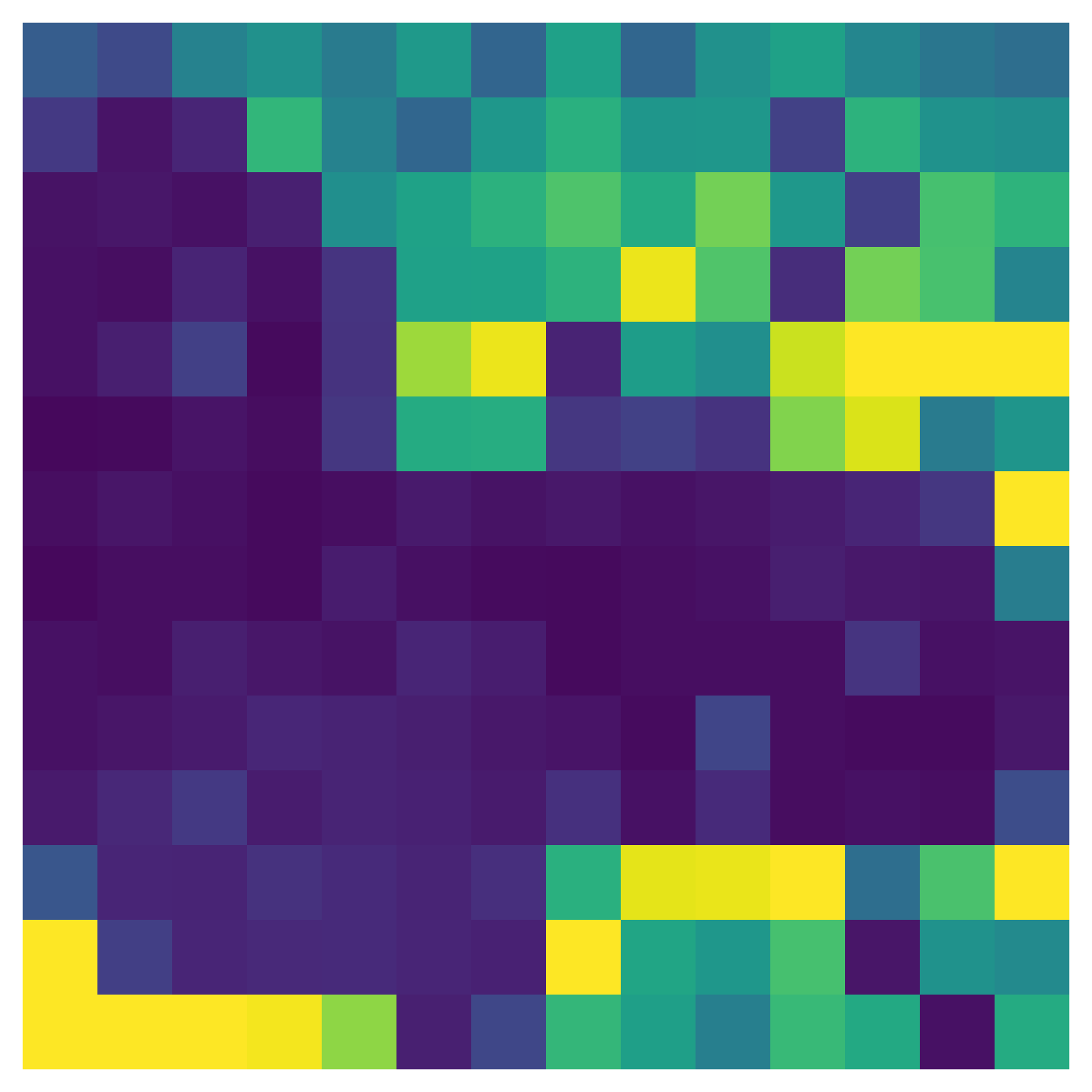}};
            \node at (6,0) {\includegraphics[width=2cm]{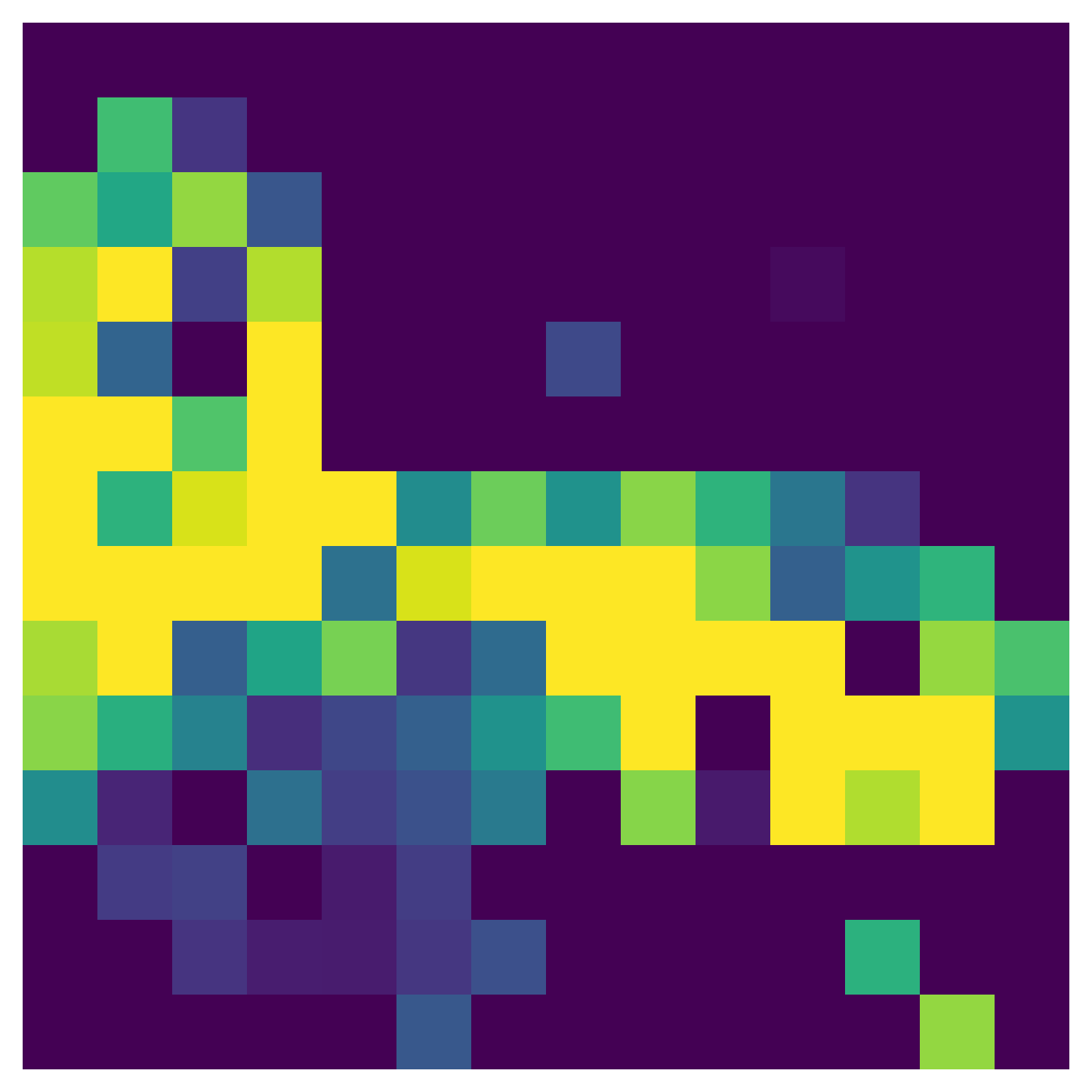}};
        \end{tikzpicture}
        \subcaption{Comparison of softmax, softmin, and inhibited attention maps ($\alpha=1.8$)}
        \label{fig:methodexamples}
    \end{subfigure}

\end{minipage}

\caption{A (a) schematic overview of the Inhibited Self-Attention, and (b) example images alongside their attention maps for softmax (conventional Self-Attention), softmin, and our Inhibited Self-Attention (ISA).} \label{fig:2}
\vspace{-4mm}
\end{figure*}
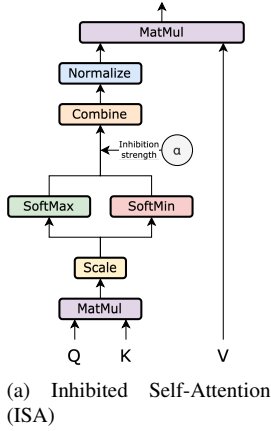
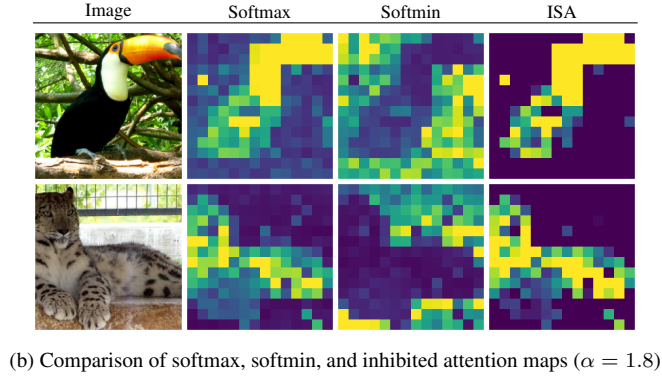

We propose a new type of self-attention mechanism for Vision Transformers that utilizes all available information in the attention, exploiting both positive and negative scores. We call it \textit{Inhibited Self-Attention} (ISA). 
The output of ISA is computed through an inhibition process, inspired by the push-pull inhibition in the visual system of the brain, which effectively uses negative attention scores usually discarded by the standard self-attention due to the softmax activation function. An overview of our proposed method is illustrated in \cref{fig:2}a.

\paragraph{(Raw) Self-Attention}
The purpose of the attention mechanism in Vision Transformers is to allow the model to learn how to dynamically focus on specific parts of the input for making predictions. This is done by calculating attention scores for each token representing an image patch, effectively determining how important each part of the image is. For the proposed inhibited self-attention, we calculate the raw attention scores as for the standard attention $\textbf{A}_{raw} = \textbf{QK}^T / \sqrt{d_\textit{k}}$,  where \textbf{Q} is the query, \textbf{K} the key, and $\textit{d}_k$ a scaling factor. 

\paragraph{Inhibited Self-Attention} 
Whereas the normal attention mechanism would only calculate the softmax of the raw attention scores, we calculate the softmin distribution as well. Given the raw attention scores \( \mathbf{A}_{\text{raw}} \in \mathbb{R}^{n \times n} \), we compute the positive attention \( \mathbf{A}^+ \)  and the negative attention \( \mathbf{A}^- \) as the softmax and softmin distributions, respectively. Their formulation is as follows:

\begin{equation}
    \mathbf{A}^+_{i,j} = \frac{e^{\mathbf{A}_{\text{raw}, i,j}}}{\sum_k e^{\mathbf{A}_{\text{raw}, i,k}}},
~\quad~
    \mathbf{A}^-_{i,j} = \frac{e^{-\mathbf{A}_{\text{raw}, i,j}}}{\sum_k e^{-\mathbf{A}_{\text{raw}, i,k}}}
\end{equation}

The softmin attention scores enable the model to capture the negative attention between patches in a complementary manner to the positive attention, assigning higher weights to elements with lower scores and effectively highlighting those patches that potentially contribute negatively to model predictions. We combine the positive and negative attention scores through an inhibition mechanism to obtain the inhibited self-attention scores:

\vspace{-0.15cm}
\begin{equation}
    \tilde{\mathbf{A}}_{i,j} = \max \left(\mathbf{A}^+_{i,j} - \alpha \mathbf{A}^-_{i,j}, \epsilon \right)
\end{equation}

\noindent where \( \alpha \) is a hyperparameter that indicates the inhibition strength and \( \epsilon = 10^{-10} \) is a small constant to ensure the function is strictly positive while maintaining numerical stability and preventing vanishing gradients.

Finally, we L1-normalize row-wise to ensure a valid probability distribution of the self-attention scores:

\vspace{-0.1cm}
\begin{equation}
    \mathbf{A}^{\text{inhib}}_{i,j} = \frac{\tilde{\mathbf{A}}_{i,j}}{\sum_k \tilde{\mathbf{A}}_{i,k}}
\end{equation}

By using the negative raw attention scores, our formulation captures both excitatory and inhibitory relationships between patches, allowing the model to infer not only what to focus on but also what to ignore. We show the difference between the standard attention mechanism using only softmax and ISA in \cref{fig:methodexamples}. The same input yields an attention map with more focused attention on the object of interest and reduced attention on the surroundings. Importantly, ISA does not merely suppress low-attention patches; it selectively increases attention on some regions while decreasing it on others, producing a reweighted map that cannot be reproduced by any pointwise (non)linear transformation of the standard attention map, since ISA relies on a competitive interaction between two distinct attention signals.

\paragraph{Deployment in ViTs}
We replace the standard self-attention mechanism in each layer of the ViT with our proposed Inhibited Self-Attention. For object detection experiments, inhibition is applied exclusively within the ViT backbone. In all experiments, $\alpha$ is set as a learnable parameter, initialized with a value of 1.

\section{Evaluation}
We demonstrate ISA’s enhanced attention selectivity through quantitative and qualitative analysis of attention maps. We compare against several baselines on image classification and object detection, evaluating performance on clean data, shortcut reliance, and out-of-distribution settings.

\subsection{Experimental Setup}
\paragraph{Datasets}
We evaluate ISA selectivity on the ImageNet-1k~\cite{deng2009imagenet} and ImageNet-Segmentation (ImageNet-S)~\cite{gao2022large} datasets. ImageNet-S, a pixel-annotated subset of ImageNet-1k, provides ground-truth segmentation maps that allow us to quantify the attention focused on objects versus background regions.

To examine shortcut vulnerability, we employ ImageNet-Watermark (ImageNet-W)~\cite{li2023whac}, which introduces Chinese character watermarks into the ImageNet-1k validation set to test sensitivity to spurious correlations.

Generalization to out-of-distribution samples is evaluated using ImageNet-Renditions (ImageNet-R)~\cite{hendrycks2021many}, while background reliance is assessed with the Waterbirds dataset~\cite{sagawa2019distributionally}. We further evaluate robustness to image corruptions on ImageNet-C~\cite{hendrycks2019benchmarking} and ImageNet-$\bar{\text{C}}$~\cite{cbar}.

Finally, to demonstrate the advantages of ISA beyond image classification, we conduct object detection experiments on the COCO dataset~\cite{cocodataset}.

\paragraph{Evaluation Metrics}
We analyze the focus of ViT attention on class-relevant regions, namely on the areas belonging to the objects of interest, using the \emph{Attention-on-Objects} metric that we propose. We elaborate on this metric in \cref{par:att}. 

The performance of the models on clean data is measured by computing the classification accuracy and cross-entropy loss on the test set. For evaluation on ImageNet-W, we report the \textit{IN-W Gap} which is the accuracy degradation compared to ImageNet-1k, and the \textit{Carton Gap} that corresponds to the increase in the carton class accuracy from ImageNet-1k to ImageNet-W \cite{li2023whac}. For corrupted data we report the \textit{mean Corruption Error (mCE)} and \textit{relative mean Corruption Error (rmCE)} \cite{hendrycks2019benchmarking}, and normalize the results of each model with inhibition with respect to those of its respective baseline without inhibition. For object detection we evaluate the models on the mean average precision (mAP) and \textit{Attention-on-Detected-Objects} (see \cref{sec:od}).

\paragraph{Baselines}
Considering that our proposed Inhibited Self-Attention is a drop-in replacement for the self-attention, we train our ViTs from scratch. We take ViT-S/16 as our baseline, which established a strong ImageNet-1k benchmark without the need of computationally expensive pre-training~\cite{beyer2022better}, together with the Differential Transformer~\cite{ye2024differentialtransformer}, which we adapt for ViTs (DIFF-ViT). To demonstrate the stability of the proposed mechanism, we also train a DeiT-S model~\cite{touvron_training_2021}, with and without ISA. For object detection, we use our pre-trained ViT-S as backbone with a DINO detection head \cite{zhangdino}. Models incorporating ISA are denoted as `ISA-ViT', `ISA-DeiT', and `ISA-DINO', corresponding to the ViT, DeiT, and DINO architectures, respectively.

\paragraph{Implementation}
We trained all models on the ImageNet-1k dataset from scratch. For the ViT-S/16, DIFF-ViT-S/16, and ISA-ViT-S/16, we adopted the training settings outlined in \cite{beyer2022better}, with 90 training epochs and RandAugment augmentation~\cite{cubuk2020randaugment}. Consistent with \cite{beyer2022better}, we employed global average pooling for final classification instead of a \texttt{CLS} token\footnote{For visualization, we also trained the models with a \texttt{CLS} token.}.
For DeiT and ISA-DeiT, we followed the official DeiT settings, training for 300 epochs with straightforward augmentations, including grayscale, solarization, Gaussian blur, and color jitter. For knowledge distillation, we used the pre-trained RegNetY-16GF~\cite{radosavovic2020designing} image classification model as the teacher model. We opted for knowledge distillation with soft targets to help the student ViT model better capture inter-class relationships. Fine-tuning for downstream tasks was performed without knowledge distillation. Instead, we applied linear probing on the concatenation of the \texttt{CLS} and \texttt{DIST} tokens. For DINO, we trained for 12 epochs and followed the configuration settings of \cite{zhangdino}. With a simple convolutional feature pyramid network we extracted 3 feature maps from the output of the ViT backbone as input for the DINO head. The batch size was set to 256 for classification and 2 for object detection. 
All training was conducted on a single node with 4 NVIDIA A100 GPUs. The code and hyperparameters used for training were kept consistent between the proposed models and respective baselines to ensure a fair comparison. Seed training was enabled for reproducible results.

\subsection{Enhanced Attention Selectivity}
We visualize the model attention using CheferCam~\cite{chefer2021generic, chefer1}, which produces more accurate visualizations compared to raw attention, roll-out \cite{abnar2020quantifying} or GradCAM \cite{selvaraju2017grad}, and investigate how ISA influences the final attention. 

\paragraph{Attention-on-Objects (AoO). } \label{par:att}
We develop the Attention-on-Object (AoO) metric to measure attention focus, defined as the ratio of attention on object regions to that on the background. Unlike the Pointing Game~\cite{li2018tell}, which considers only the single highest attention point, AoO accounts for all attention mass, and uses segmentation masks from ImageNet-S to delineate object and background regions, rather than the coarser boxes employed by the Energy-based Pointing Game~\cite{wang2020score}. The AoO ratio ($A_{\text{object}}$) is calculated as:

\begin{equation}
A_{\text{object}} = \frac{\sum_{i,j} \mathcal{M}_{i,j}A_{i,j}}{\sum_{i,j} A_{i,j}},
\end{equation}

\noindent with \( A_{i,j} \) denoting the attention value at pixel \( (i,j) \) of the interpolated attention map, and $\mathcal{M}_{i,j} \in \{0,1\}$ a binary mask indicating whether pixel $(i,j)$ belongs to the object of interest. As we are not only interested in the amount of attention on the object but the strength of the attention, we incrementally threshold the attention and obtain the thresholded AoO ($A_{\text{object}}^{\text{thresh}}$) as follows:

\begin{equation}
A^{thresh}_{\text{object}} = 
\frac{\sum_{i,j} M_{i,j} A^{t}_{i,j}}{\sum_{i,j} A^{t}_{i,j}},
\quad
A^{t}_{i,j} =
\begin{cases}
A_{i,j}, & A_{i,j} \ge T_\text{val} \\
0, & \text{otherwise}
\end{cases},
\label{threshold_formula}
\end{equation}

\noindent where \( T_{\text{val}} = \frac{T}{100} \times A_\text{max} \) is the threshold value, with \( T \) being the threshold percentage (in the range of [0, 99]), and \( A_{\text{max}} \) the normalized maximum value in the attention map. This thresholding further distinguishes AoO from the Pointing Game~\cite{li2018tell}, which considers only the single strongest attention point, and the Energy-based Pointing Game~\cite{wang2020score}, which does not account for attention strength, providing a more complete picture of how attention strength is spatially distributed across object and background regions.

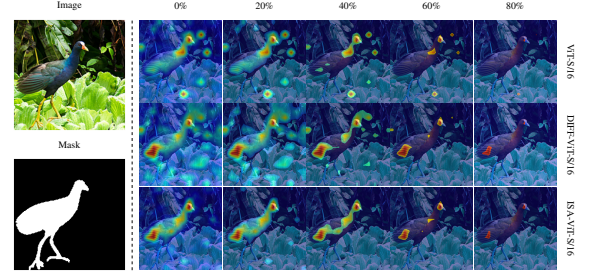
\begin{figure}[t]
\centering
\begin{adjustbox}{width=0.9\linewidth}

\begin{tikzpicture}

\node[font=\normalsize] at (0, 10) {Image};
\node[font=\normalsize] at (0, 5) {Mask};

\node at (0,7.5) {\includegraphics[width=4cm]{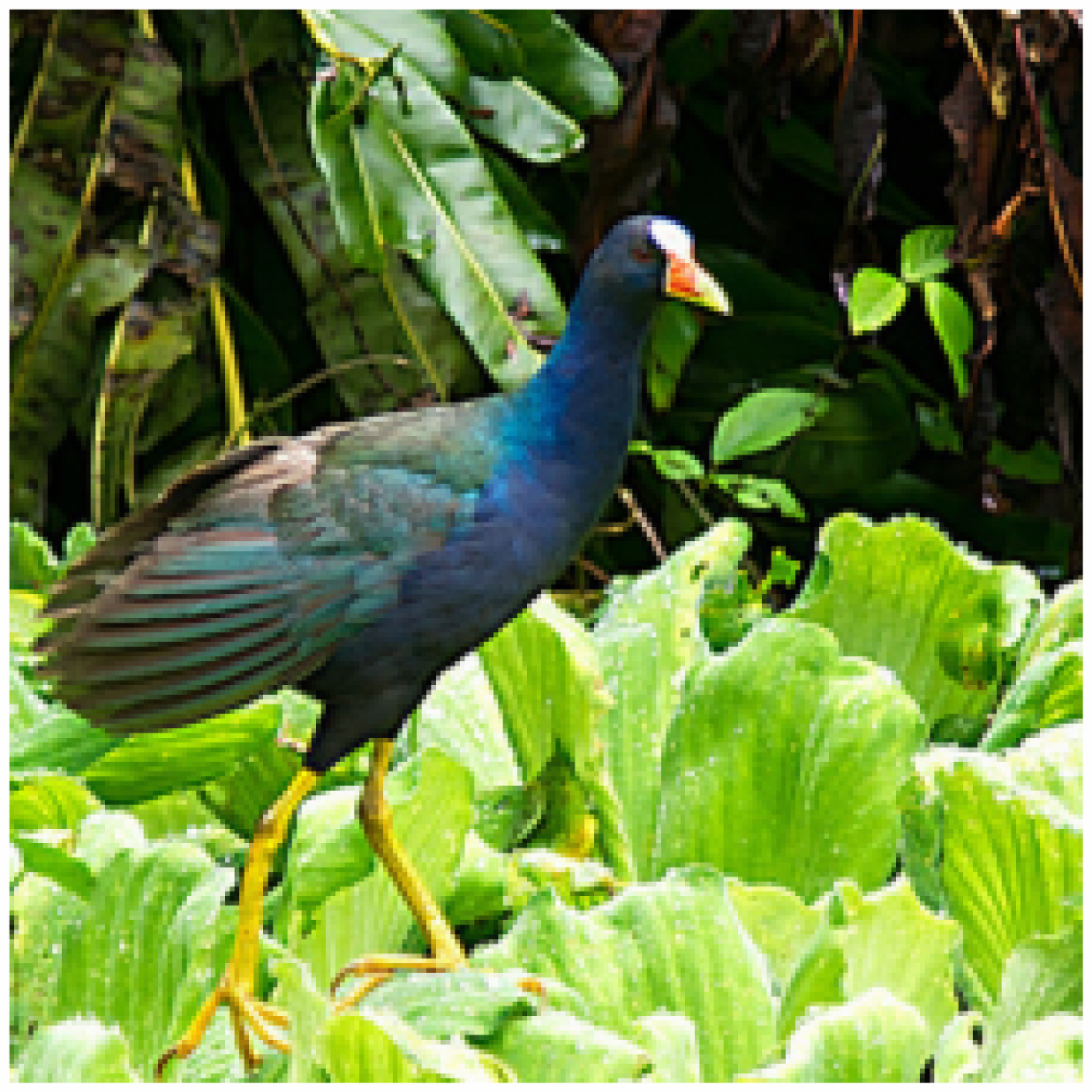}};
\node at (0,2.5) {\includegraphics[width=4cm]{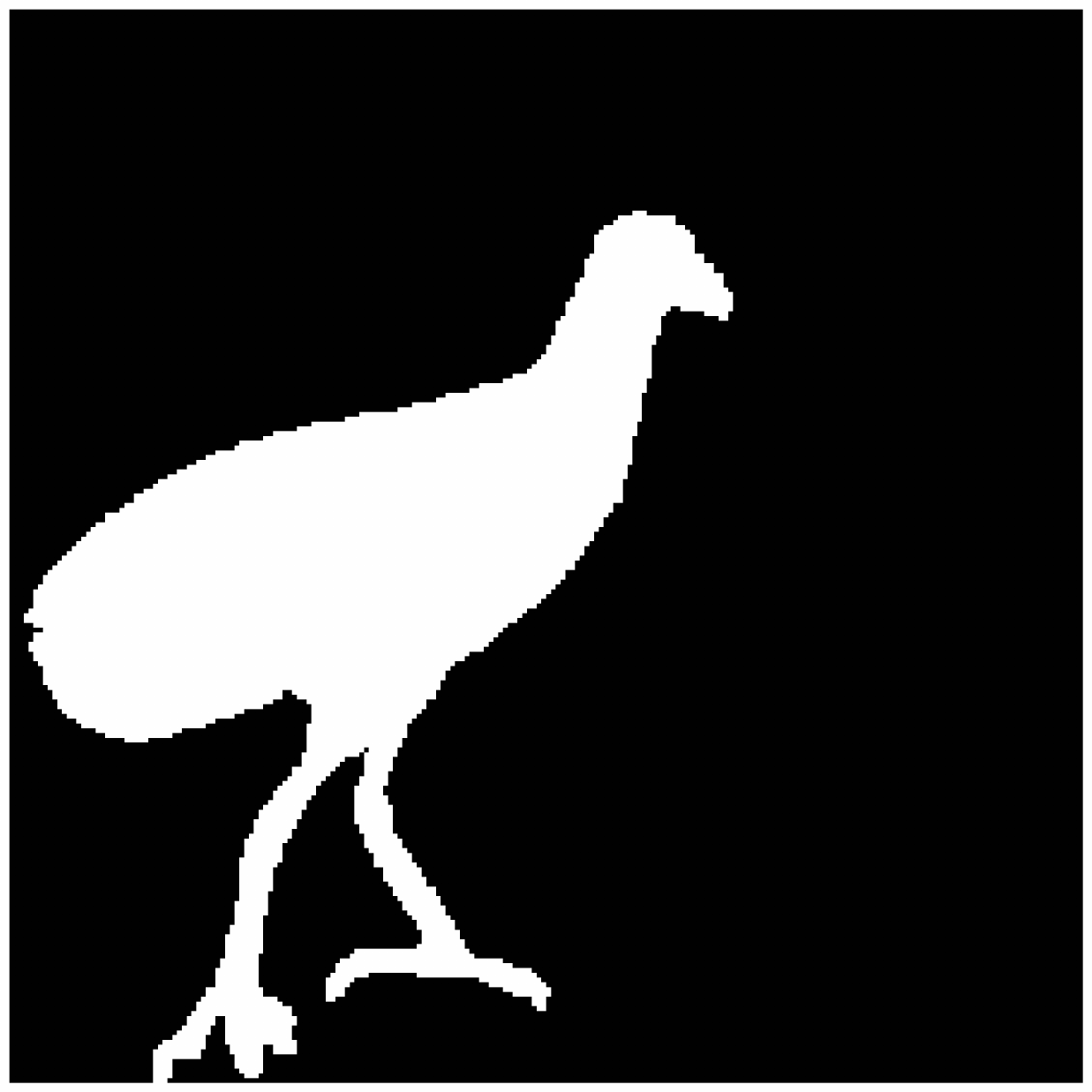}};

\node at (2.25, 5) {\tikz \draw[dashed] (0,0) -- (0,9);};

\node[font=\normalsize] at (4,10) {0\%};
\node[font=\normalsize] at (7,10) {20\%};
\node[font=\normalsize] at (10,10) {40\%};
\node[font=\normalsize] at (13,10) {60\%};
\node[font=\normalsize] at (16,10) {80\%};

\node[font=\normalsize, rotate=270] at (18,8) {ViT-S/16};
\node[font=\normalsize, rotate=270] at (18,5) {DIFF-ViT-S/16};
\node[font=\normalsize, rotate=270] at (18,2) {ISA-ViT-S/16};

\node at (4,8)  {\includegraphics[width=3cm]{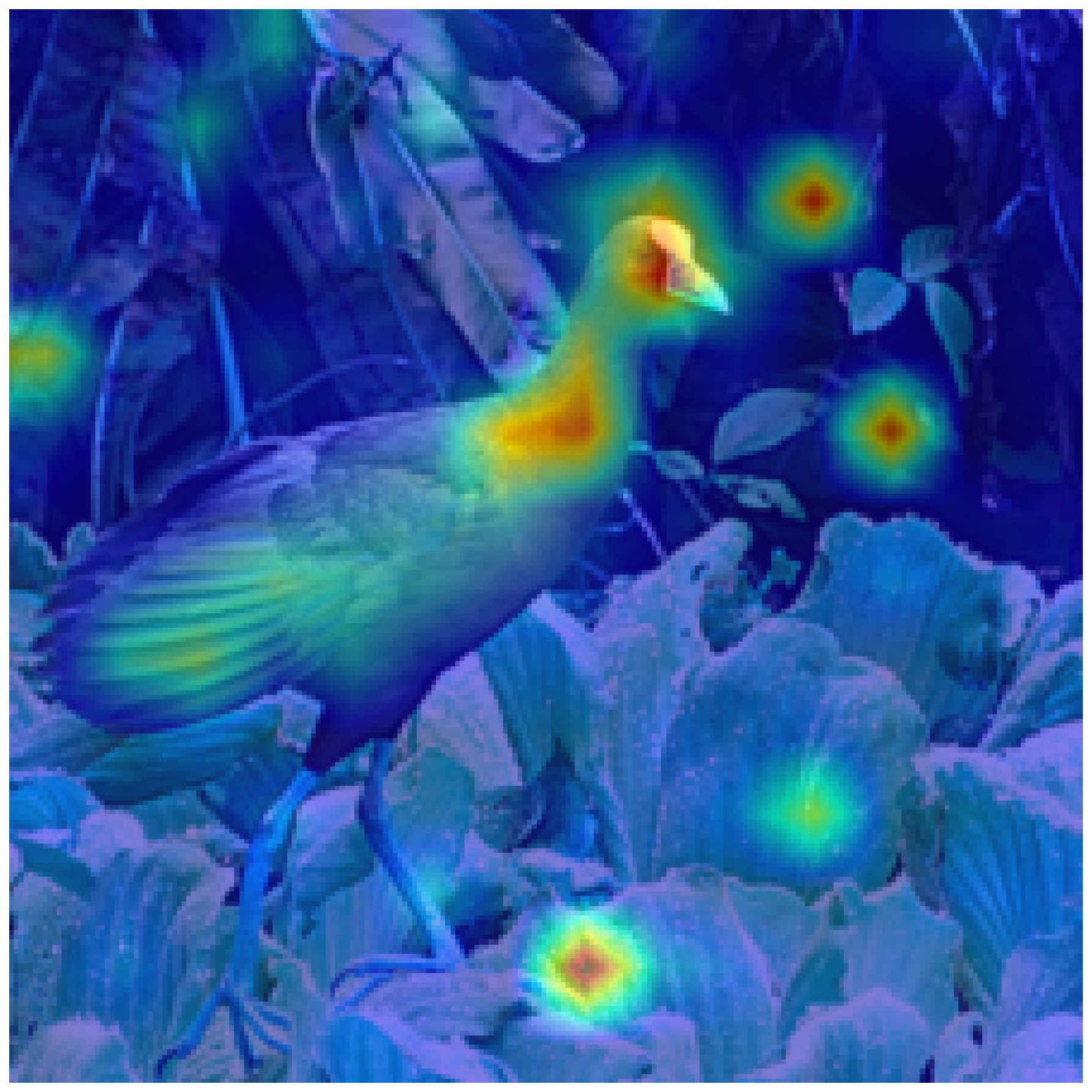}};
\node at (7,8)  {\includegraphics[width=3cm]{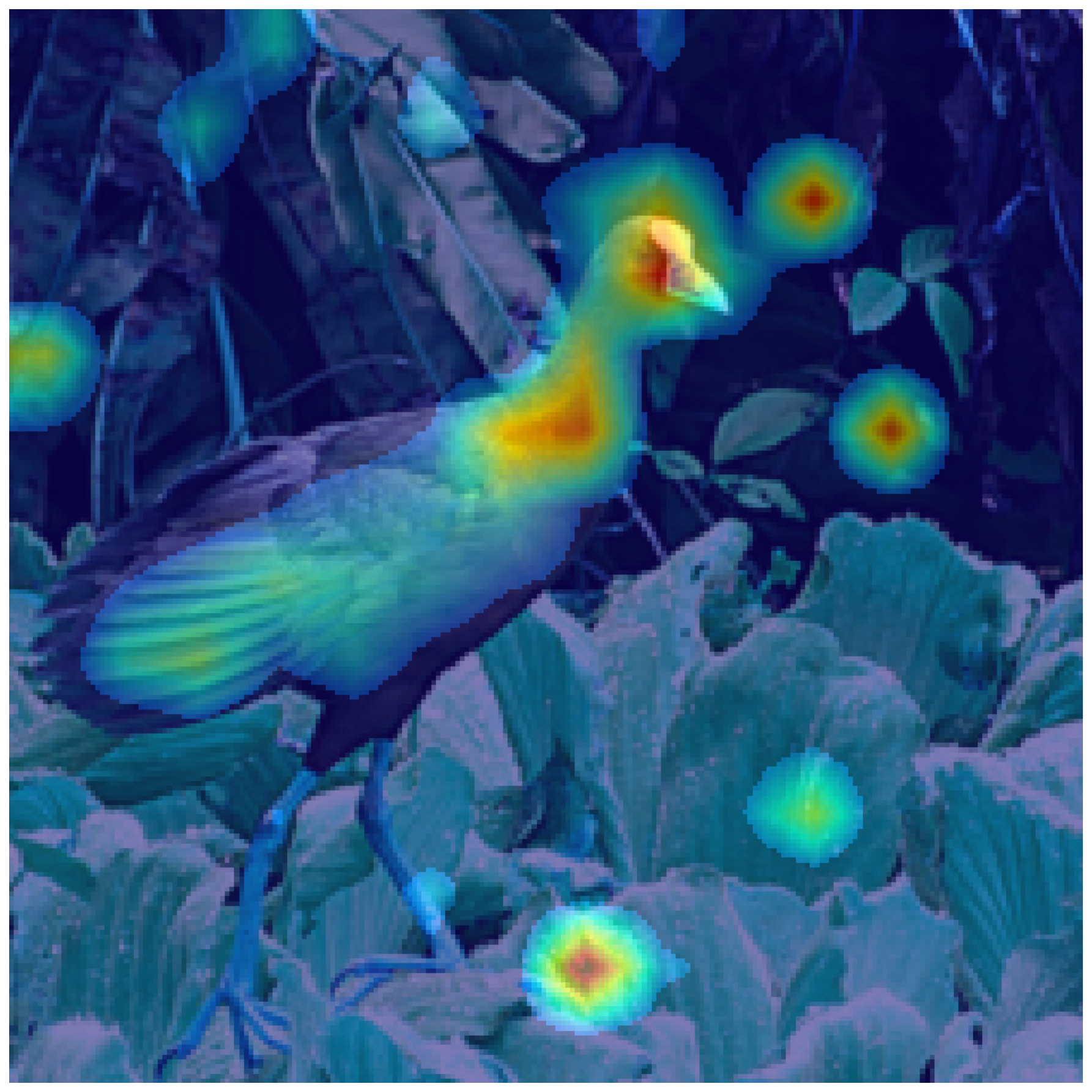}};
\node at (10,8) {\includegraphics[width=3cm]{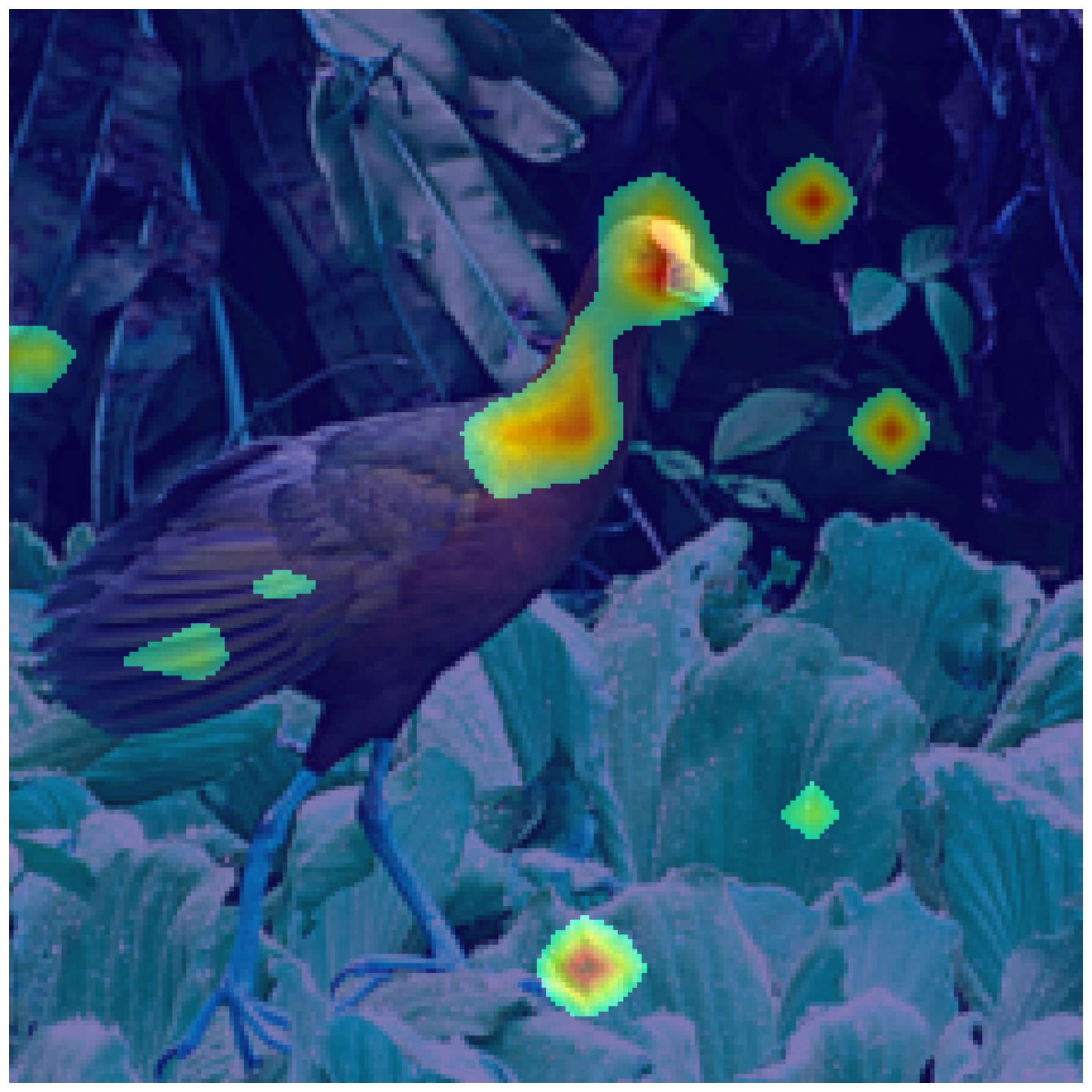}};
\node at (13,8) {\includegraphics[width=3cm]{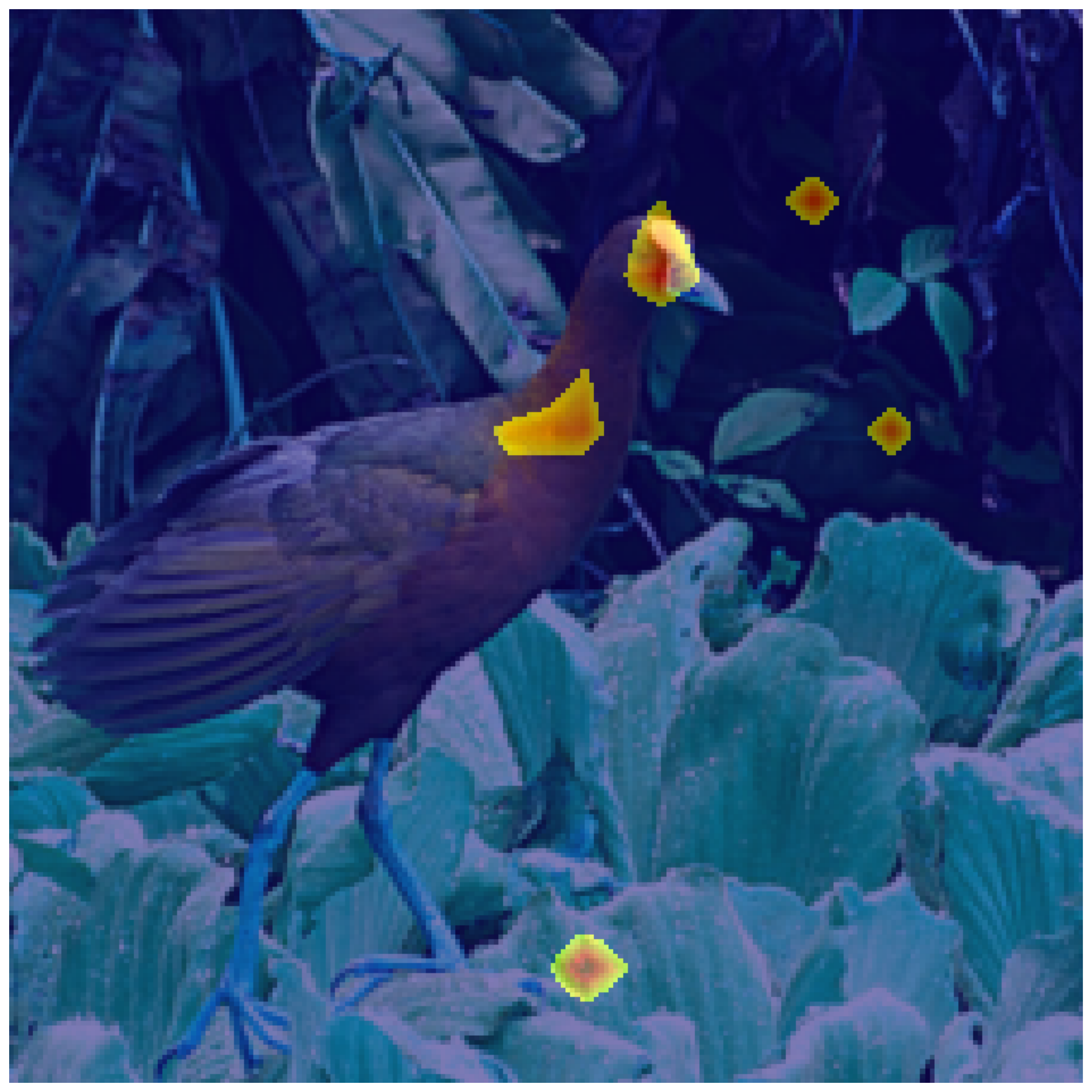}};
\node at (16,8) {\includegraphics[width=3cm]{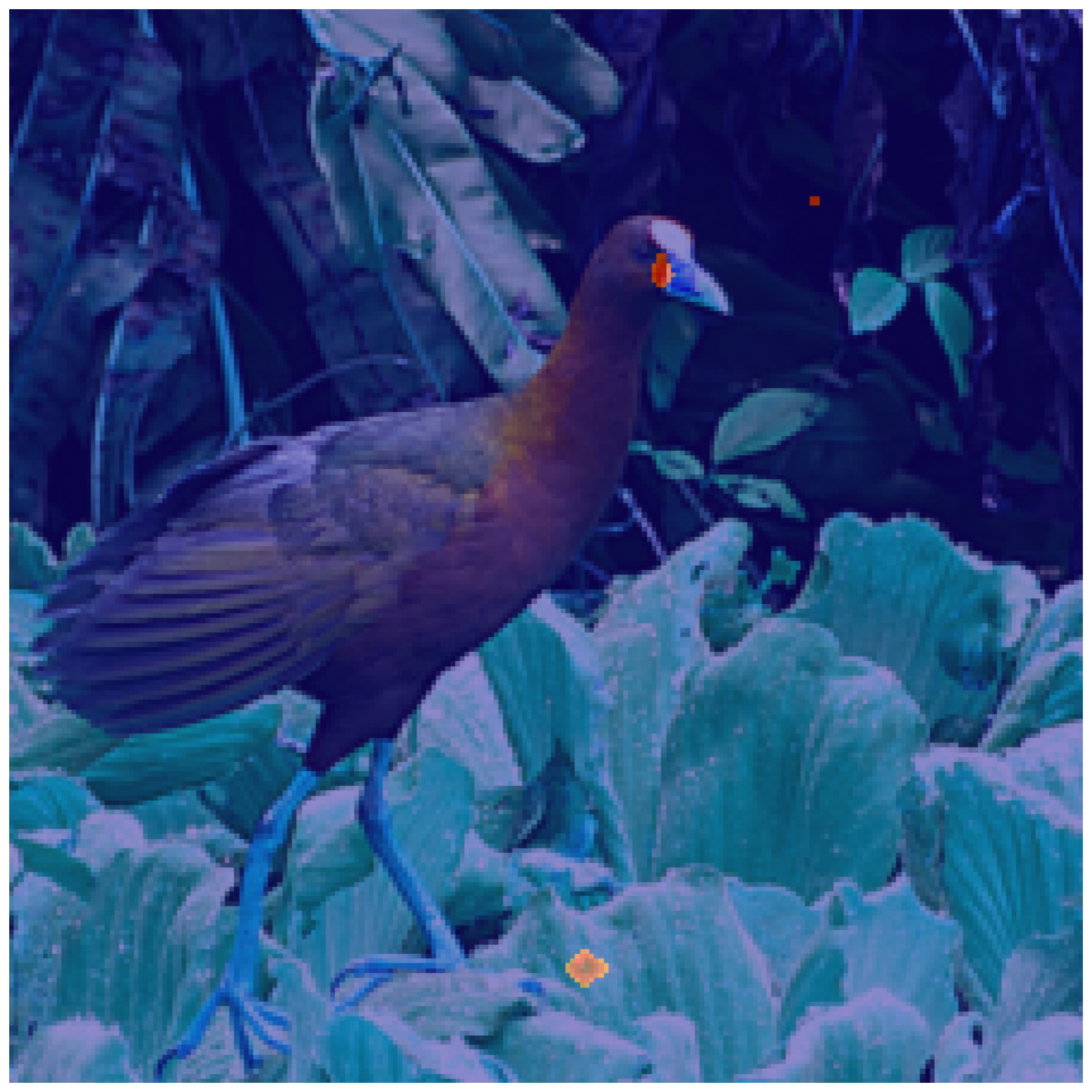}};

\node at (4,5)  {\includegraphics[width=3cm]{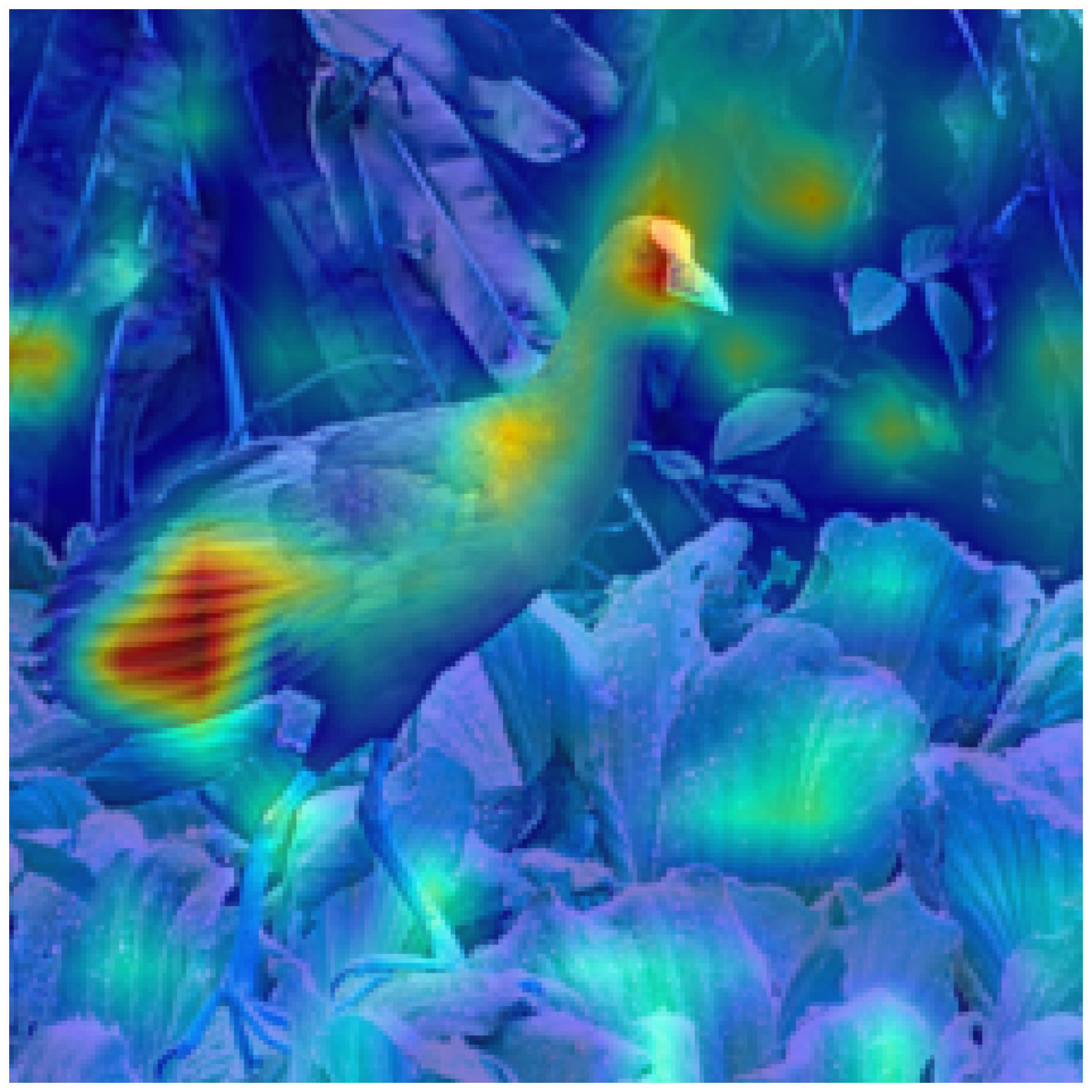}};
\node at (7,5)  {\includegraphics[width=3cm]{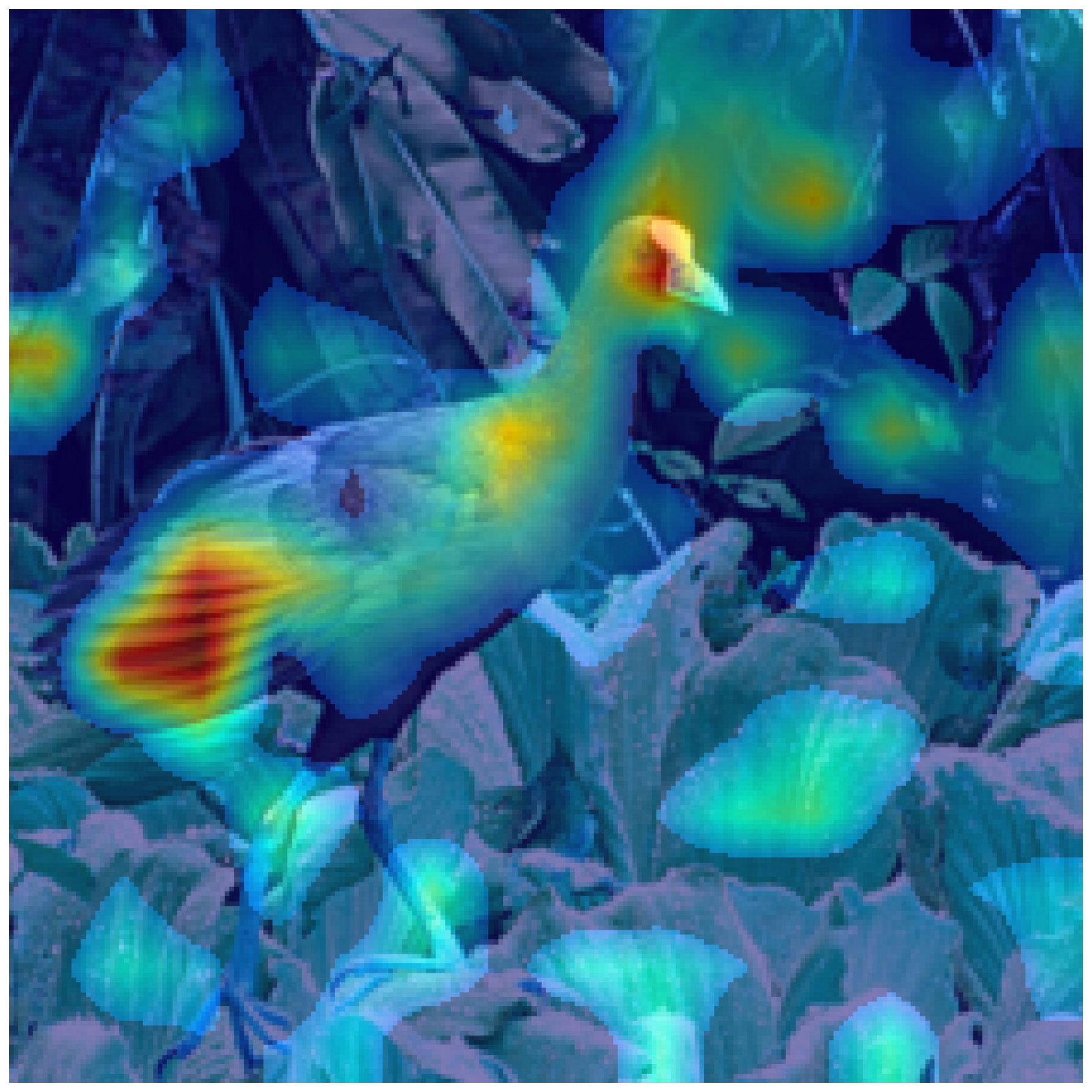}};
\node at (10,5) {\includegraphics[width=3cm]{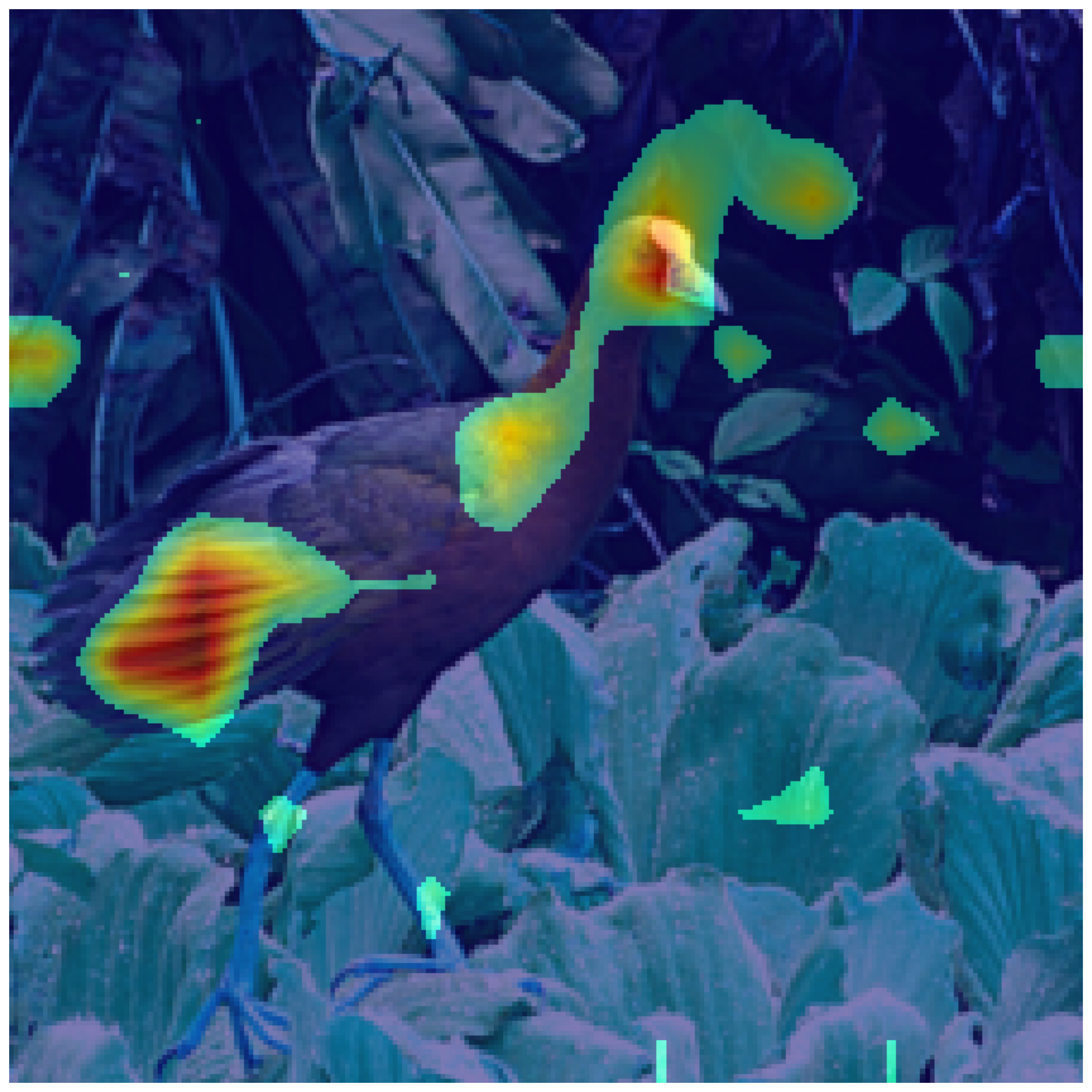}};
\node at (13,5) {\includegraphics[width=3cm]{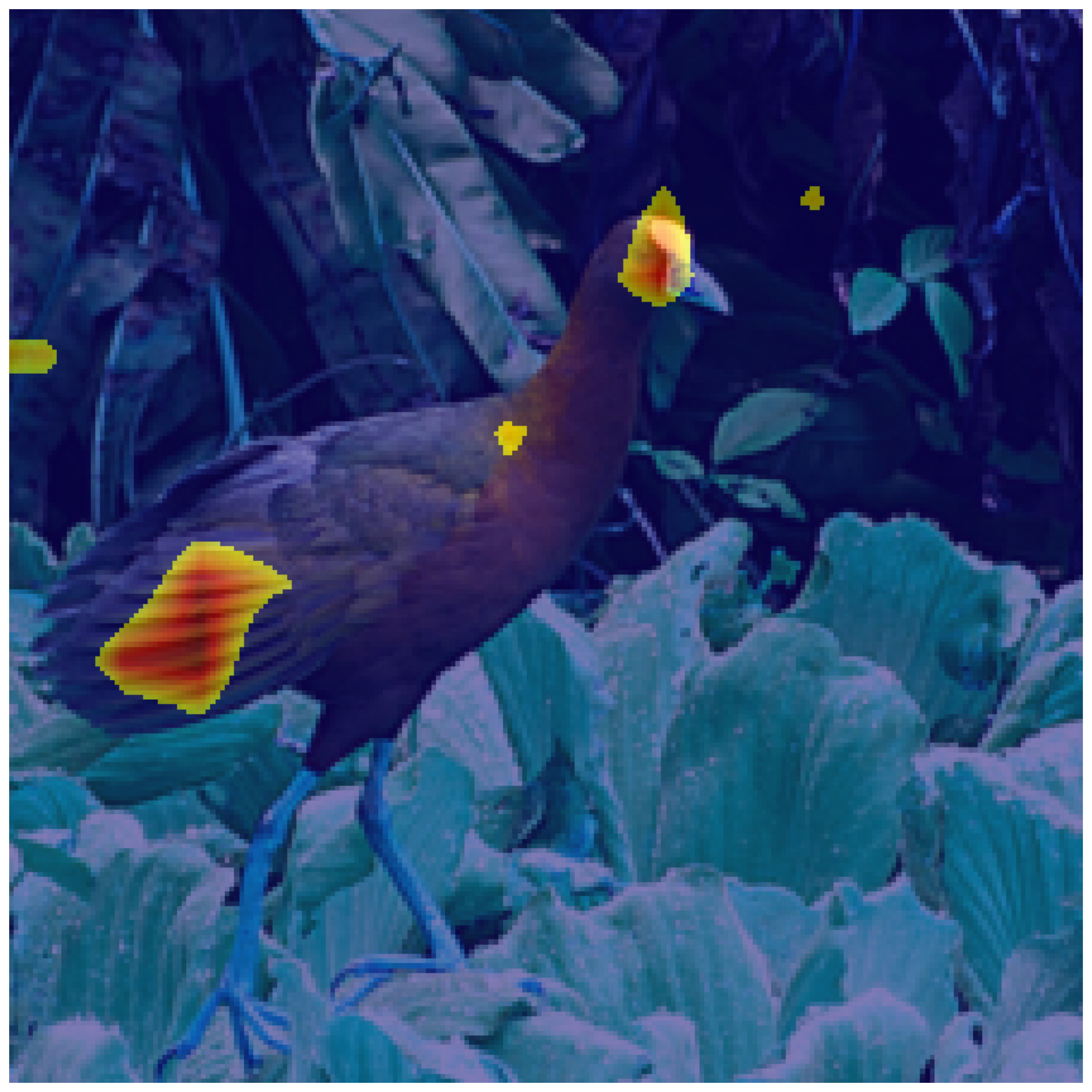}};
\node at (16,5) {\includegraphics[width=3cm]{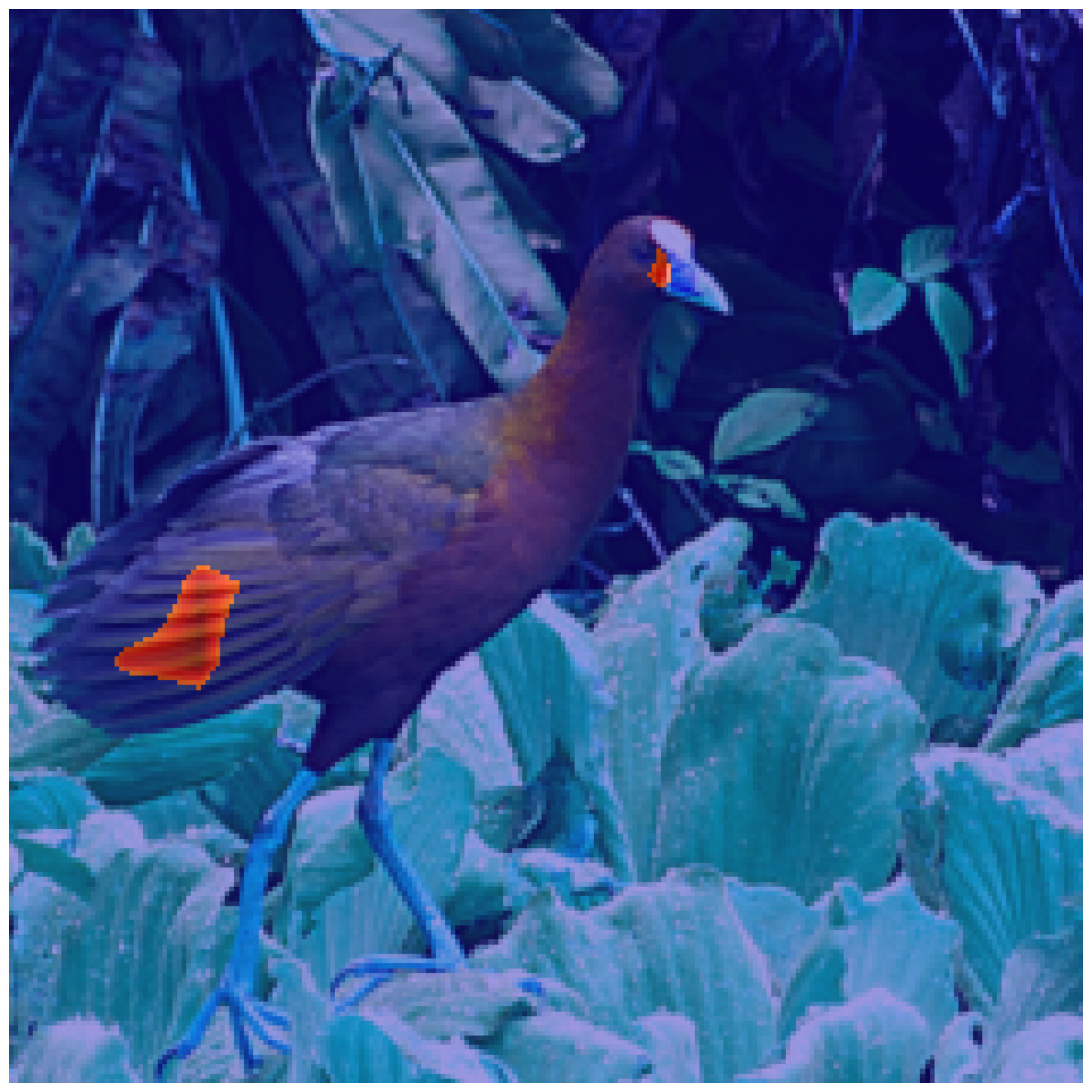}};

\node at (4,2)  {\includegraphics[width=3cm]{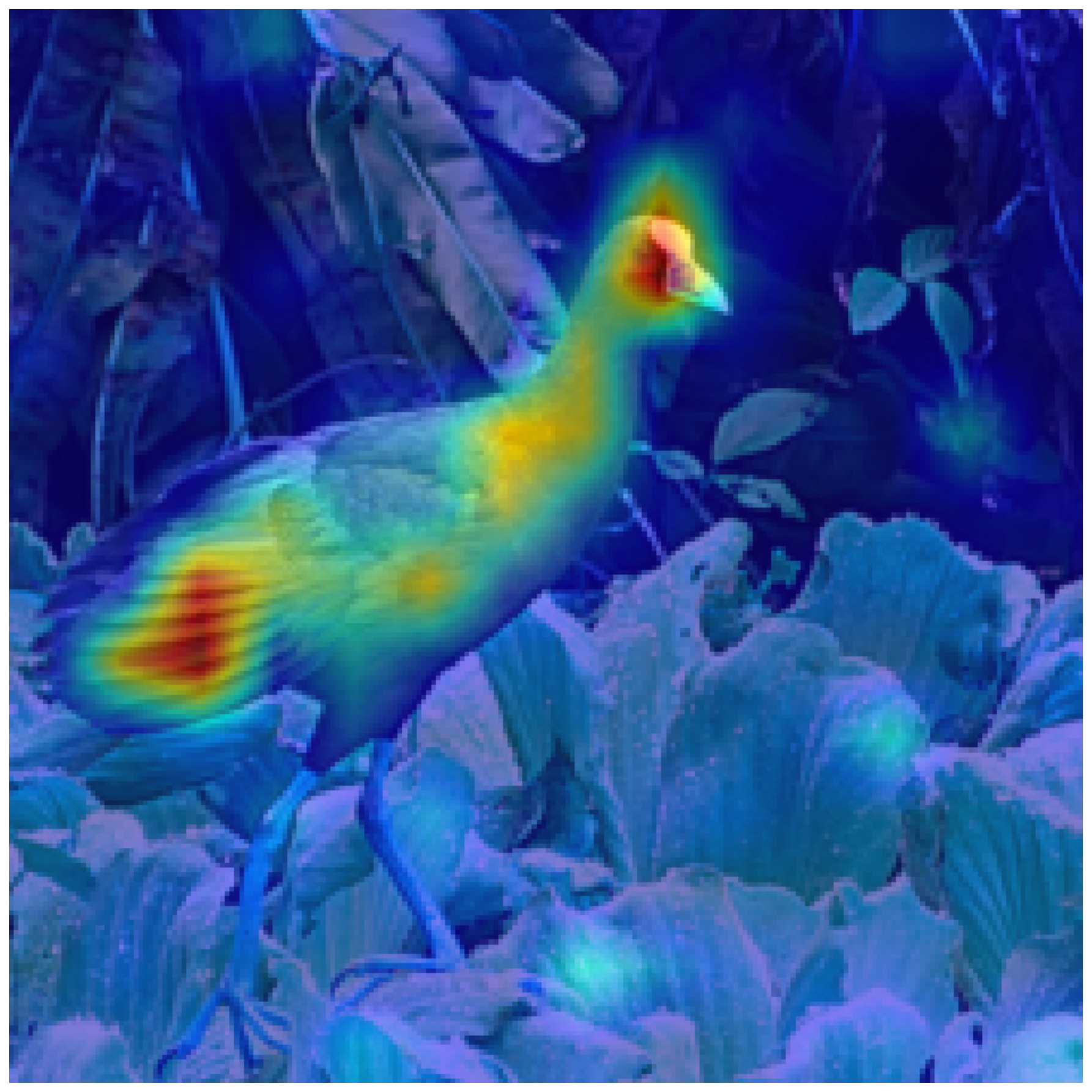}};
\node at (7,2)  {\includegraphics[width=3cm]{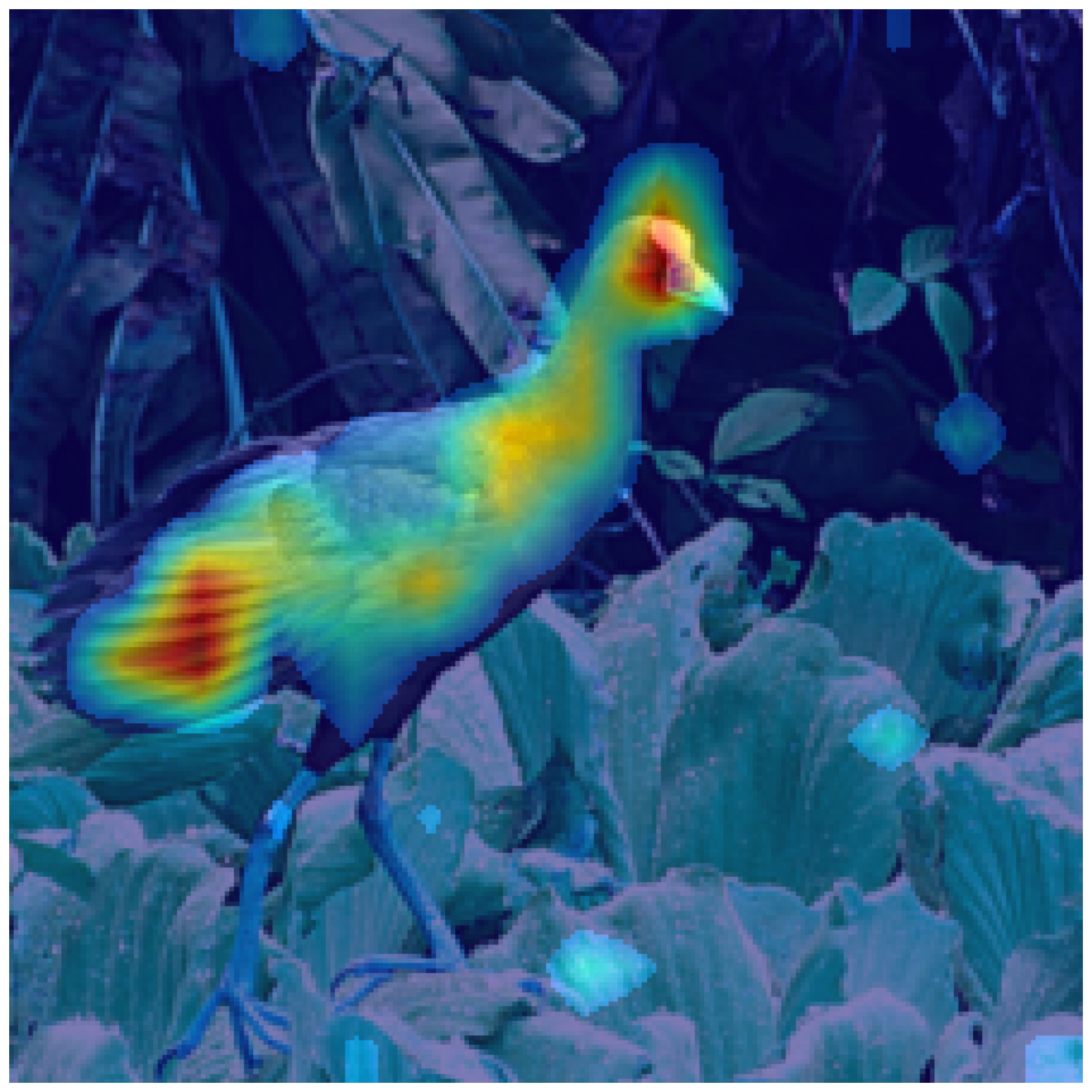}};
\node at (10,2) {\includegraphics[width=3cm]{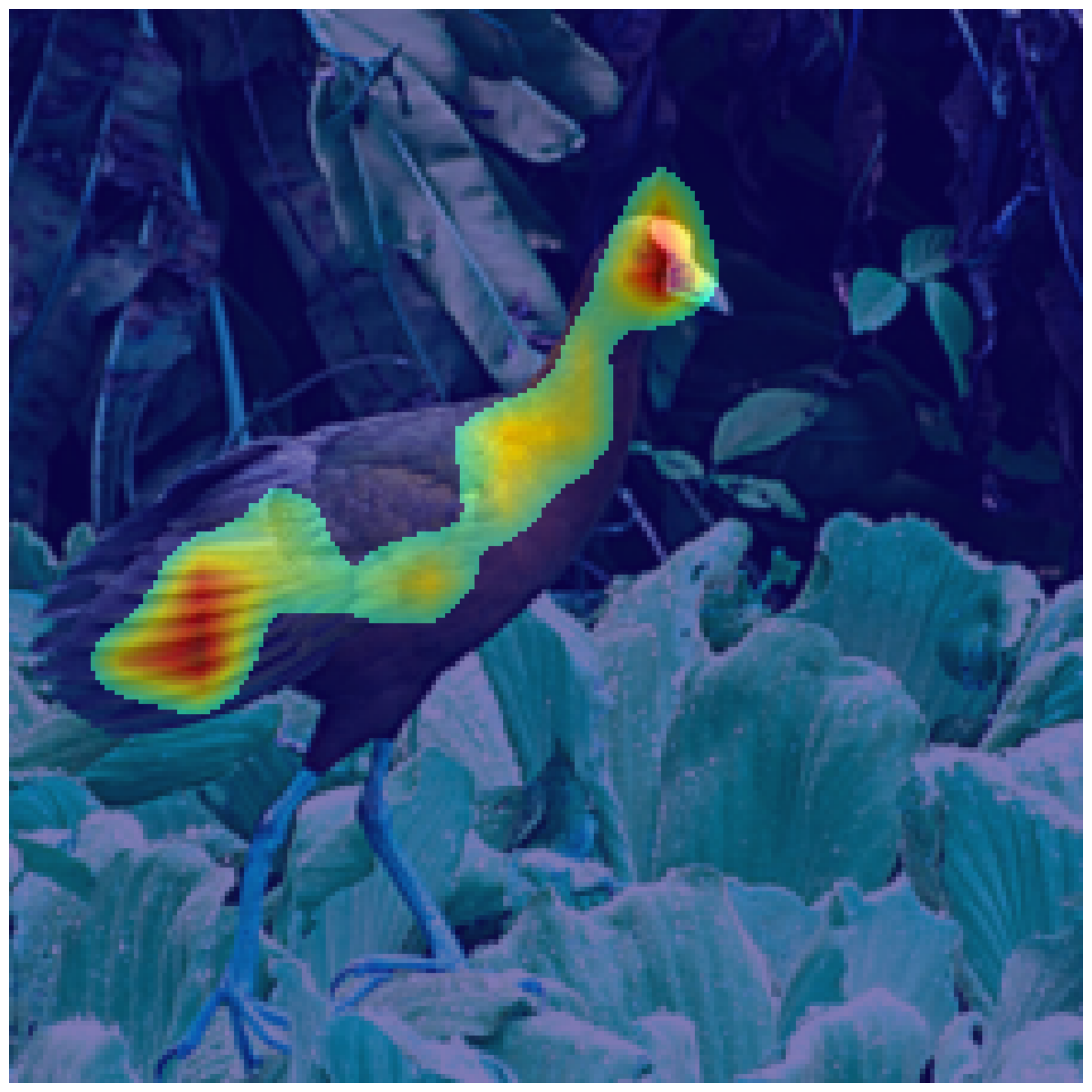}};
\node at (13,2) {\includegraphics[width=3cm]{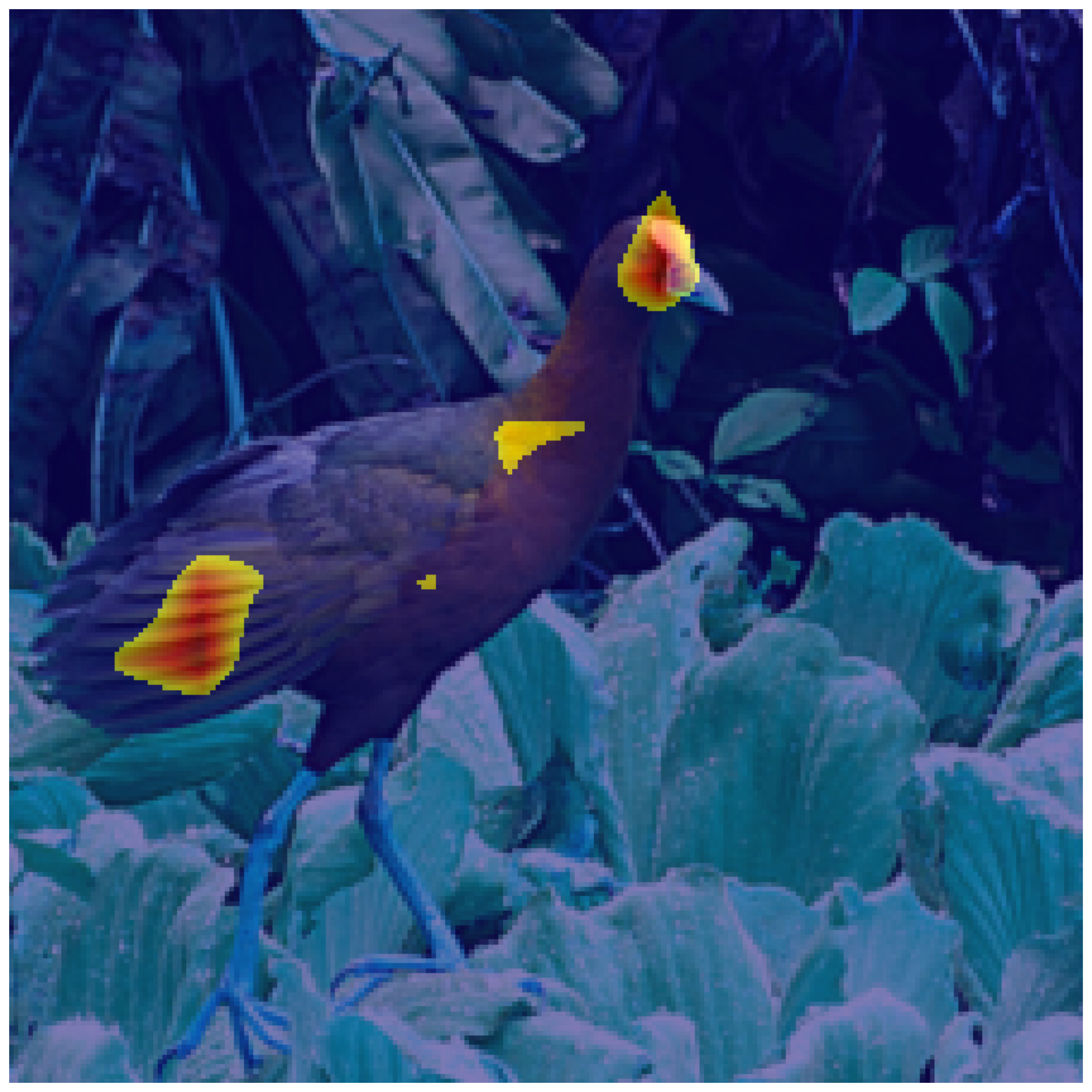}};
\node at (16,2) {\includegraphics[width=3cm]{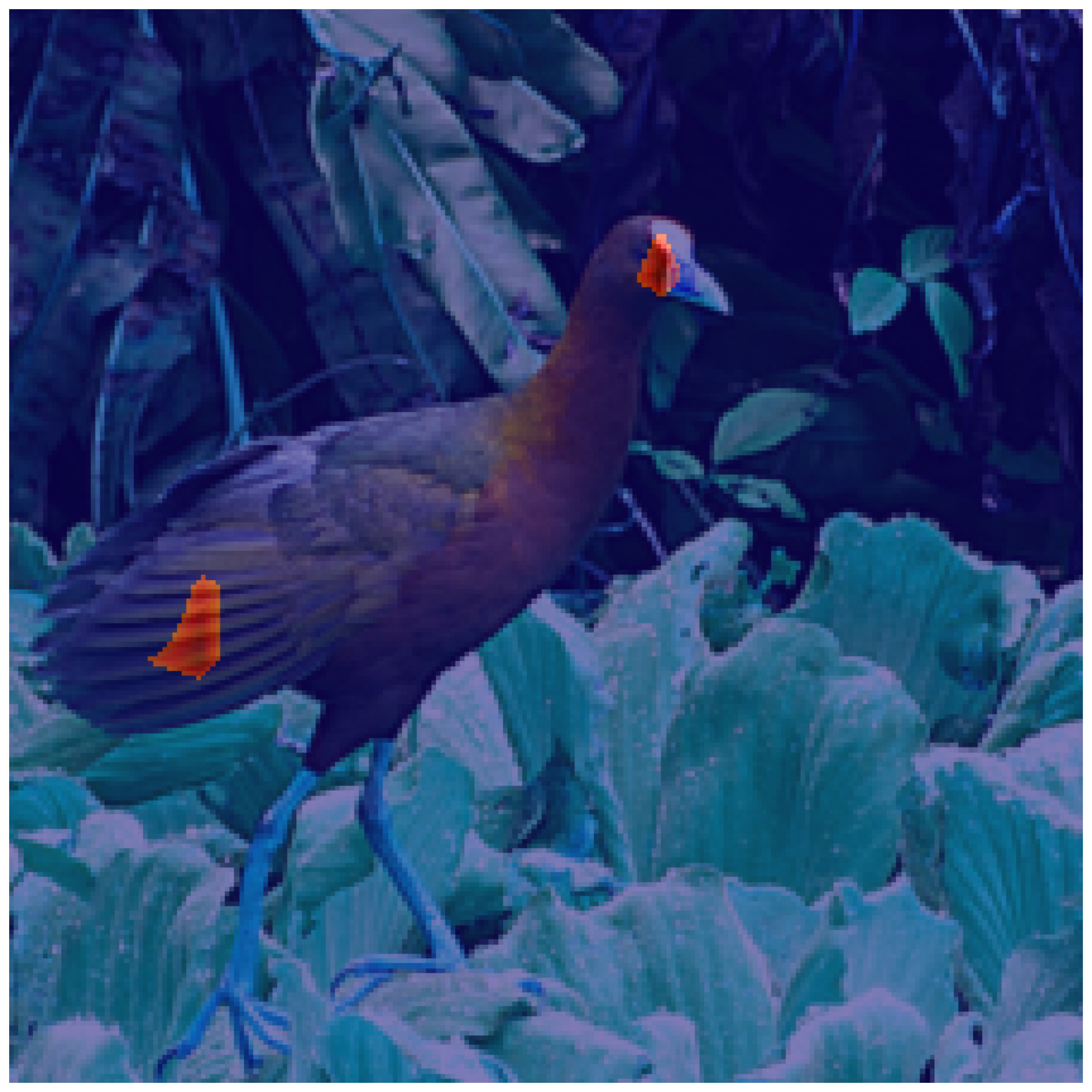}};

\end{tikzpicture}

\end{adjustbox}

\caption{Original image and ground-truth mask (left). Thresholded attention maps at different percentages of the maximum attention value for ViT-S/16, DIFF-ViT-S/16, and ISA-ViT-S/16 (right).}

\label{fig:thresh_viz}

\vspace{-0.1cm}
\end{figure}

\begin{figure*}[t!]
  \centering
  \begin{subfigure}{0.3\textwidth}
    \centering
    \includegraphics[width=\linewidth]{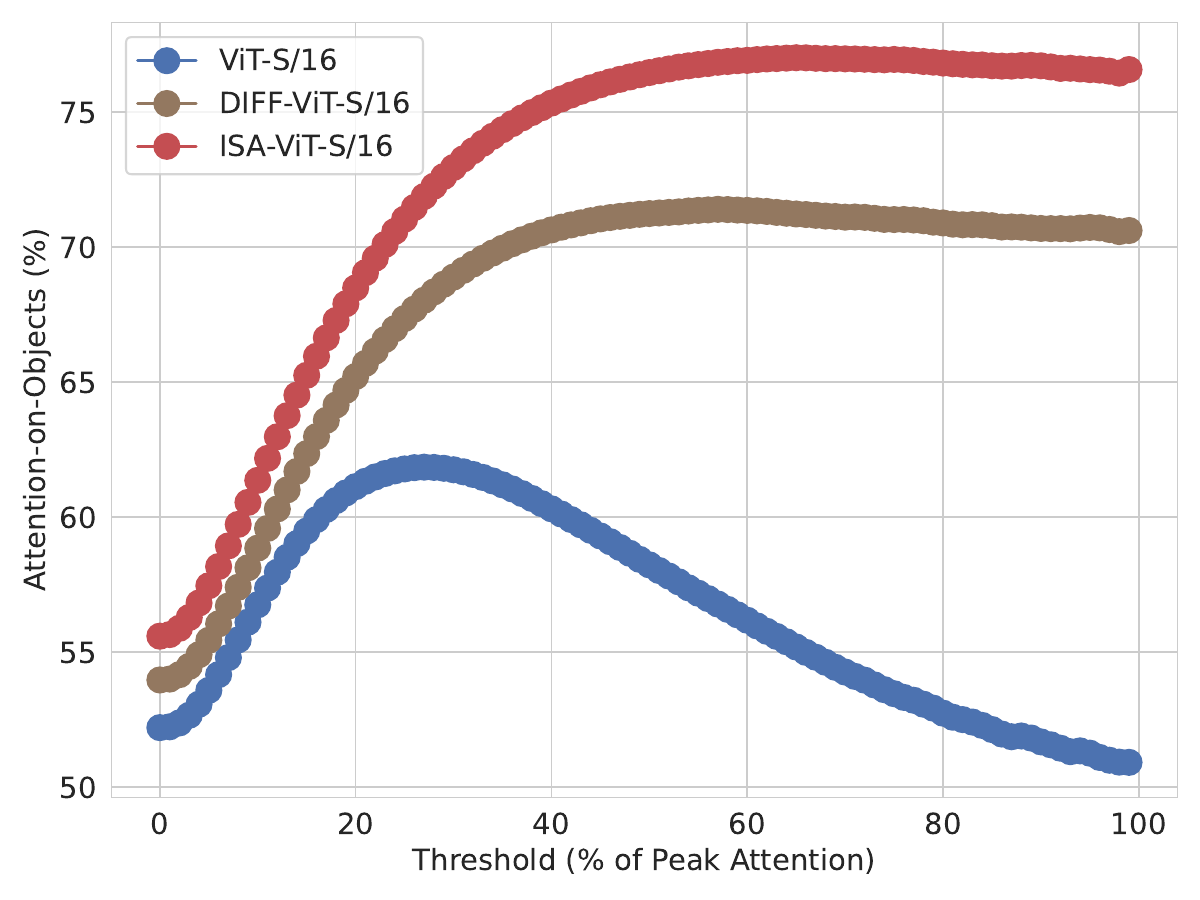}
    \subcaption{}
    \label{fig:threshold}
  \end{subfigure}
  \begin{subfigure}{0.3\textwidth}
    \centering
    \includegraphics[width=\linewidth]{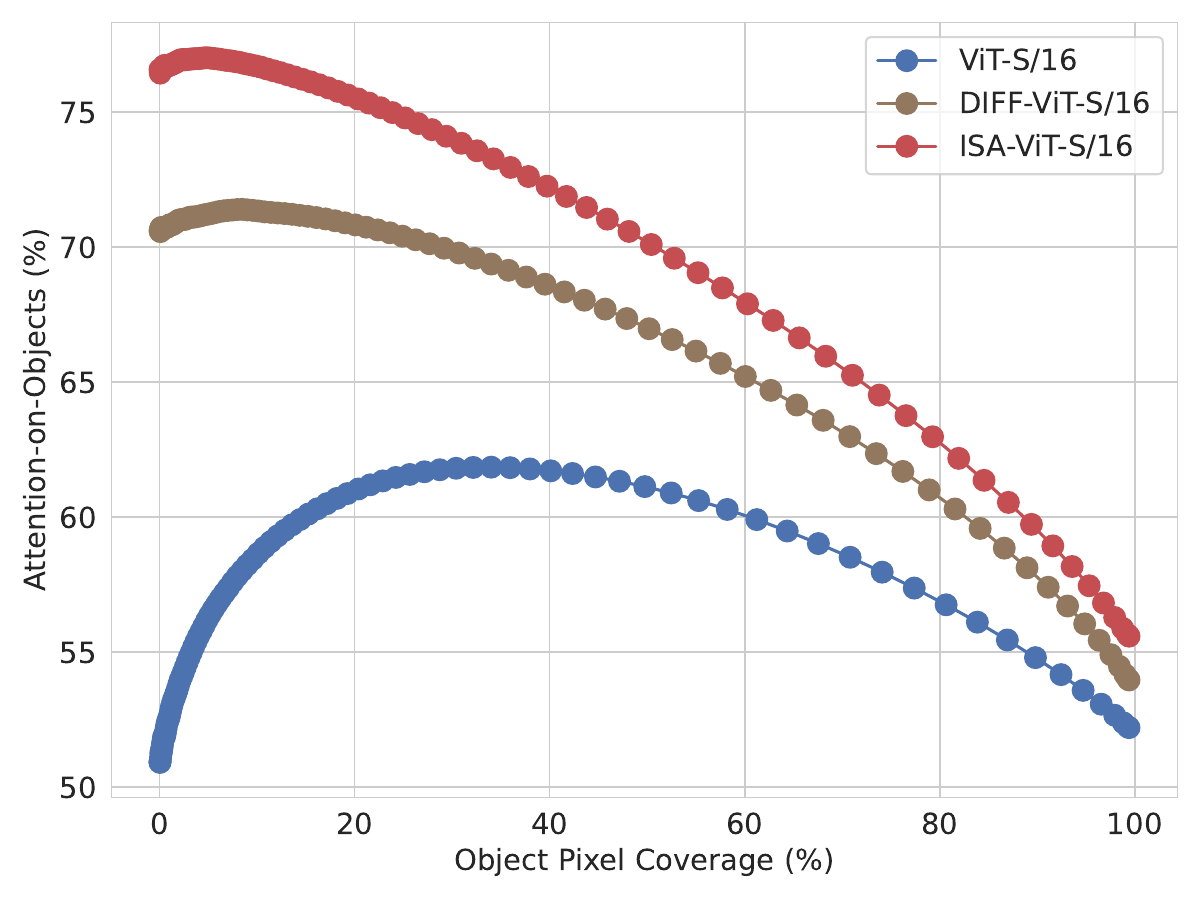}

    \subcaption{}
    \label{fig:pr}
  \end{subfigure}
  \begin{subfigure}{0.3\textwidth}
    \centering
    \includegraphics[width=\linewidth]{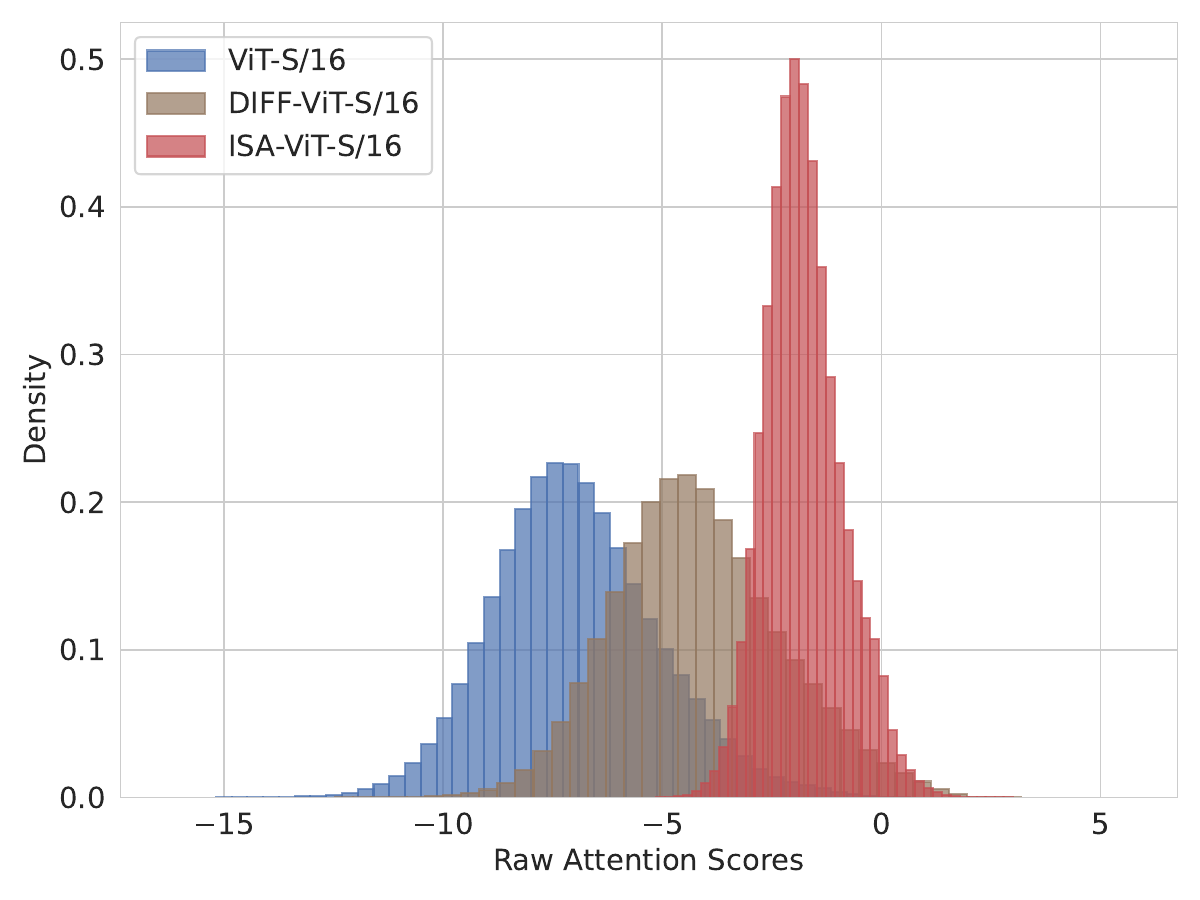}
    \subcaption{}
    \label{fig:distributionshift}
  \end{subfigure}
  \caption{(a) Ratio of attention on the object of interest (AoO) for threshold levels of the normalized peak attention in the range $[0,99]$ on ImageNet-S \cite{gao2022large}. The ratio was calculated at each threshold for all ±12.5k validation images with available segmentation maps. (b) AoO precision-recall curve computed for the same validation images, with higher thresholds retaining only the most confidently attended tokens. (c) Raw attention score distributions of the last layer. The plain ViT's wider distribution causes peaked, winner-takes-all attention, while ISA's narrower distribution yields more balanced, object-focused attention.}
  \vspace{-0.2cm}
\end{figure*}

An example attention map and the corresponding segmentation mask are illustrated in the left part of \cref{fig:thresh_viz}. The right part of \cref{fig:thresh_viz} shows that at higher thresholds, ISA-ViT-S/16 retains significantly more attention on the object compared to the baseline without inhibition. We computed the AoO ratio for all validation images at each threshold in the range $[0,99]$ according to~\cref{threshold_formula} and plot the average ratio at each threshold in~\cref{fig:threshold}. The figure illustrates that our inhibited self-attention leads to a higher Attention-on-Object ratio across all threshold levels. Notably, at higher threshold levels, the gap in the ratio between ISA-ViT-S/16 and ViT-S/16 exceeds 25 percentage points, highlighting that our proposed mechanism directs significantly stronger attention to the objects of interest. DIFF-ViT-S/16 lies between the two, improving over the baseline but plateauing around 71\% and trailing ISA-ViT-S/16 by roughly 5 percentage points at high thresholds, indicating less concentrated attention on objects.

We additionally analyze AoO and object coverage across the same 100 thresholds, and plot a precision-recall style curve in \cref{fig:pr}. At threshold 0 all tokens receive attention and object coverage is maximal, corresponding to the rightmost point of the curve. As the threshold increases, only the most confidently attended tokens remain, moving the curve leftward toward lower coverage. ISA dominates across the full curve, maintaining high AoO even as coverage shrinks, meaning its highest-confidence tokens consistently fall on objects. ViT's AoO peaks at intermediate coverage and degrades toward the high-confidence end, confirming that its most-attended tokens frequently fall outside object boundaries. DIFF-ViT improves over ViT but plateaus well below ISA across the full curve, with the gap widening toward the high-confidence end, showing that its output-level subtraction cannot match ISA's selectivity precisely where selectivity matters most.

\vspace{-0.15cm}
\paragraph{Attention scores per layer. }
To identify the mechanistic source of ISA's selectivity, we examine the raw attention score distribution of the last layer and visualize them in \cref{fig:distributionshift}. The plain ViT exhibits a wide score distribution, which causes softmax to produce highly peaked, winner-takes-all attention concentrated on only a few tokens. DIFF-ViT shows a similarly wide distribution, spreading attention broadly but not selectively. ISA acts as a regularizer that narrows this distribution, resulting in more balanced attention weights over a focused set of object-relevant tokens. This is further confirmed by an analysis of attention concentration metrics in Appendix A.2, showing that ISA avoids single-token fixation while concentrating its attention on object-relevant regions. Further visualizations and analysis at different layers are provided in Appendix A.1.


\paragraph{Attention Visualizations} 
In \cref{fig:attention}, we visualize the attention maps (computed according to~\cite{chefer2021generic, chefer1}) for some sample images from ImageNet-1k. ISA has a significant impact on the strength and focus of the attention on the objects of interest. Generally, the attention maps of ISA-ViT are more object focused compared to those of the baseline. After analyzing numerous maps, we found that the baseline ViT and DIFF-ViT frequently rely on shortcuts and spurious background features. This tendency is illustrated in \cref{fig:intro} and in the example image of a cat in the fifth column of \cref{fig:attention} where the baseline models make correct classifications without attending to the object. In many of these instances, ISA-ViTs did successfully attend to the objects of interest. Also for cases where the baseline ViT had reasonably good focus on the object (see examples in the first four columns of \cref{fig:attention}), our proposed method frequently reduces attention on background patches, with consequently higher attention values on the object. While ISA does not always achieve strictly higher object pixel coverage (example of the dog in the last column of \cref{fig:attention}), the baseline advantage in such cases often reflects diffuse spreading across object boundaries rather than meaningful focus on discriminative parts. Extra visualizations are provided in Appendix B.

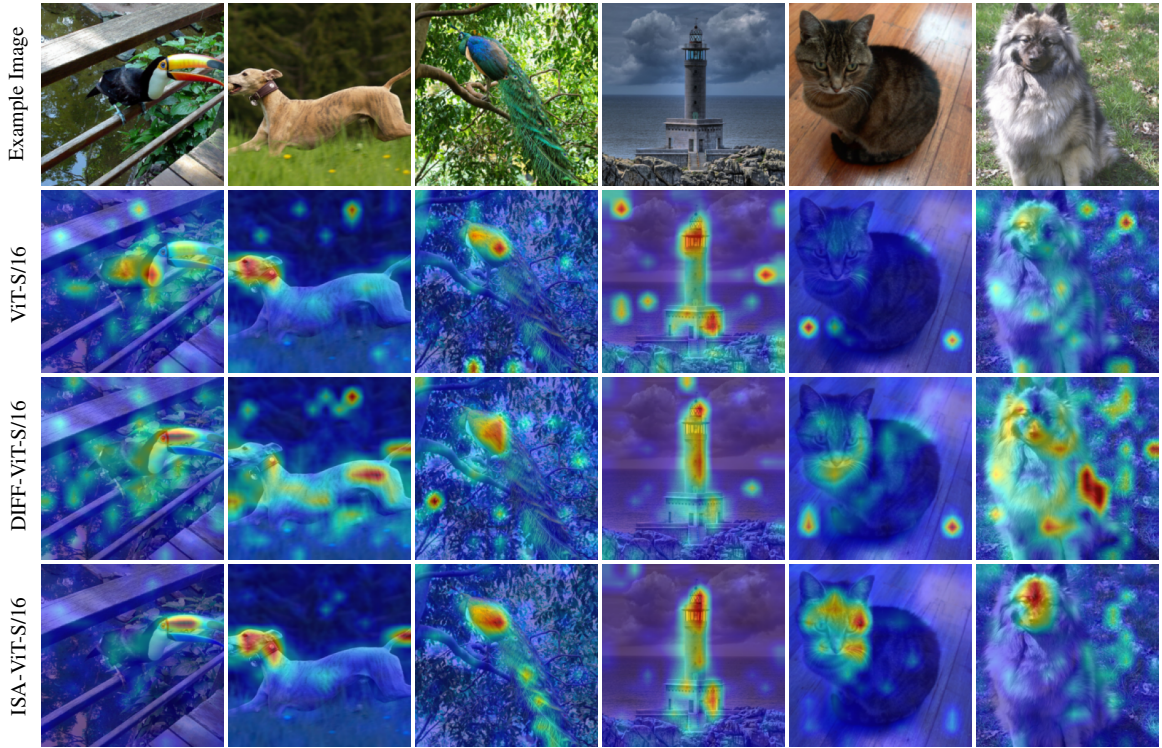
\begin{figure*}[!t]
  \vspace{-0.5cm}
  \centering
  \small
  \makebox[\columnwidth][c]{%
    \begin{tikzpicture}
    \centering
       \node at (0,0) {\includegraphics[width=0.85\linewidth]{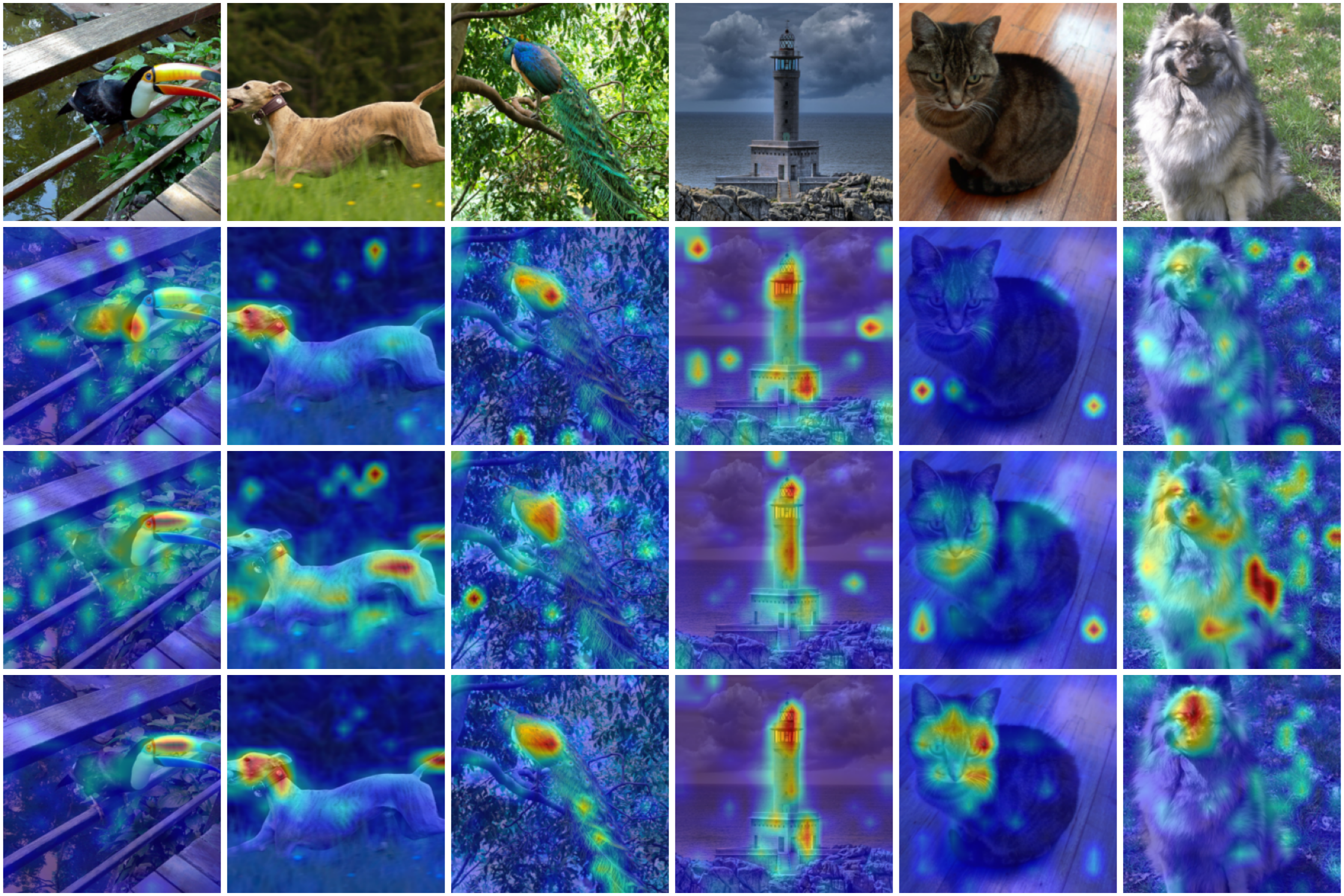}};
    \node[font=\fontsize{8}{9}\selectfont, rotate=90] at (-7.7, 3.72) {Example Image};
    \node[font=\fontsize{8}{9}\selectfont, rotate=90] at (-7.7, 1.2) {ViT-S/16};
    \node[font=\fontsize{8}{9}\selectfont, rotate=90] at (-7.7, -1.2) {DIFF-ViT-S/16};
    \node[font=\fontsize{8}{9}\selectfont, rotate=90] at (-7.7, -3.72) {ISA-ViT-S/16};
    \end{tikzpicture}
  }

    \caption{Example attention maps. ISA-ViT shows more focused attention, with higher values on the object of interest and reduced distraction from background cues.}
  \label{fig:attention}
  \vspace{-0.25cm}
\end{figure*}

\subsection{Results: Image Classification}
We evaluated ViT-S/16, DIFF-ViT-S/16, and ISA-ViT-S/16 on ImageNet-1k and assessed whether higher-quality attention maps lead to improved performance. Additionally, we tested shortcut reliance and model generalization to OOD samples by evaluating the models on ImageNet-W and ImageNet-R. Since ImageNet-R contains only 200 ImageNet classes, we fine-tuned our models on ImageNet-200 using linear probing, following the same training setup but with 10\% of the training epochs. To highlight the stability of ISA in other architectures, we also tested DeiT-S and ISA-DeiT-S. We ran each experiment three times and report the average for all metrics. 

\vspace{-0.2cm}
\paragraph{Accuracy, Shortcuts and  OOD Evaluation} In~\cref{tab:results}, we report the results on ImageNet-1K, ImageNet-W and ImageNet-R. All results are averaged over three independent runs. ISA-ViTs improve robustness and OOD performance while maintaining competitive accuracy on ImageNet-1K. For ImageNet-W, we report the IN-W Gap and Carton Gap, which quantify the top-1 accuracy degradation and the class-wise accuracy improvement for the carton class, respectively, relative to ImageNet-1k. Lower IN-W Gap indicates that the model is more robust to the presence of the watermark across all classes, while a smaller Carton Gap suggests decreased reliance on the watermark shortcut for the carton class. For all models with ISA, we improve both ImageNet-W metrics, demonstrating reduced shortcut reliance. Similarly, on ImageNet-R, models incorporating ISA consistently achieve higher accuracy on OOD samples. Unlike DIFF-ViT, ISA introduces negligible overhead in model size and FLOPs compared to the ViT-S/16 baseline; detailed comparisons are given in Appendix D.

\begin{table}[!t]
    \centering
    \scriptsize
    \setlength\tabcolsep{-1.5pt}
    \setlength\extrarowheight{2pt}
    \caption{Numerical evaluation on ImageNet-1k, ImageNet-W, and ImageNet-R. Acc. denotes the top-1 accuracy on the test set, CE Loss the cross-entropy loss on the test set, IN-W gap the accuracy degradation compared to ImageNet-1k, and the Carton Gap is the increase in class-accuracy for the carton class.}
    \label{tab:results}

    \begin{tabular*}{0.95\columnwidth}{@{\extracolsep{\fill}} l c c c c c}
    \toprule
    \multirow{2}{*}{Model} & \multicolumn{2}{c}{\textbf{ImageNet-1k}} & \multicolumn{2}{c}{\textbf{ImageNet-W}} & \textbf{ImageNet-R}\\  \cmidrule{2-3}\cmidrule{4-5}\cmidrule{6-6}
    & Acc. (\%) $\uparrow$ & CE Loss $\downarrow$  & IN-W Gap $\uparrow$ & Carton Gap $\downarrow$ & Acc. (\%) $\uparrow$  \\
    \midrule
    ViT-S/16 & 78.9 ± 0.1 & 0.890 ± 0.0 & -10.0 ± 0.8 & 28.7 ± 6.4 & 39.4 ± 0.1\\
    DIFF-ViT-S/16 & \textbf{79.1 ± 0.0} & \textbf{0.868 ± 0.0}  & -10.0 ± 1.5  & 31.3 ± 1.2 & 39.2 ± 0.1 \\ 
    ISA-ViT-S/16 & 79.0 ± 0.1 & 0.889 ± 0.0  &  \textbf{-9.3 ± 0.4} & \textbf{27.0 ± 3.1} & \textbf{40.4 ± 0.1}\\ \cmidrule[0.1pt]{2-6}
    DeiT-S & 79.9 ± 0.0 & 0.956 ± 0.0  & -8.25 ± 1.0 & 28.7 ± 3.1  & 42.2 ± 0.3 \\
    ISA-DeiT-S & \textbf{80.1 ± 0.1} & \textbf{0.943 ± 0.0} &  \textbf{-7.84 ± 0.3} & \textbf{23.0 ± 3.1} & \textbf{42.9 ± 0.4} \\

    \bottomrule

    \end{tabular*}
    \vspace{-0.3cm}
\end{table}

\vspace{-0.2cm}
\paragraph{Reliance on Background}
Given the increased object-focused attention in ISA-ViT, we investigate whether this corresponds to a reduced reliance on background features for classification of objects of interest. We use the Waterbirds dataset~\cite{sagawa2019distributionally}, which consists of photographs of various landbird and waterbird species. To introduce spurious correlations, the authors altered the backgrounds in 5\% of the training images, placing waterbirds on land backgrounds and landbirds on water backgrounds. This results in four distinct groups: landbirds on land (0), landbirds on water (1), waterbirds on land (2), and waterbirds on water (3). At test time, landbirds and waterbirds are equally distributed across land and water backgrounds. Despite high overall accuracy, models may still underperform on specific subgroups due to reliance on spurious correlations~\cite{sagawa2019distributionally}. We therefore use worst-group accuracy (i.e., waterbirds on land) as a key metric to assess susceptibility to confounding factors.

We fine-tuned ImageNet-1k pre-trained models for 30 epochs using linear probing. Each experiment was repeated three times, and we report the means in \cref{tab:wb}.

ISA-ViT yields higher class-wise accuracy on both under-represented groups during training with non-natural backgrounds (groups 1 and 2), indicating reduced reliance on spurious background features. Our method improves the worst-group accuracy by more than 2 percentage points on average with respect to ViT-S/16. The reported average accuracy is the weighted accuracy on the test set with the weights corresponding to the proportion of each group in the skewed training dataset, for which our model also outperforms the baseline.

\begin{table}[!t]
  \caption{Classification results on the Waterbirds dataset (WB). Acc. is the weighted average classification accuracy on the test set. The worst group accuracy is underlined.}
  \label{tab:wb}
  \scriptsize
  \newcommand{\gr}[0]{\cellcolor{green!25}}
  \centering
  \begin{tabularx}{0.85\columnwidth}{l C C C C C}
    \toprule
     \multirow{2}{*}{Model} & \multirow{2}{*}{Acc. (\%)} & \multicolumn{4}{c}{\bfseries Group Acc. (\%)} \\
    \cmidrule(lr){3-6}
    && 0 & 1 & 2 & 3\\
    \cmidrule(r){1-6}
     ViT-S/16 & 89.5 & 94.2 & 87.0 & \underline{81.6}  & 89.6 \\
     DIFF-ViT-S/16 & 89.3 & 93.3 & 87.2 & \underline{82.9} &	89.5 \\
     ISA-ViT-S/16 & 90.5 & 94.6 & 88.8 & \textbf{\underline{83.8}} & 89.2 \\
     \cmidrule(lr){2-6}
     DeiT-S & 91.0 & 95.9 & 88.6 & \underline{81.5}  & 91.8 \\
     ISA-DeiT-S & 90.9 & 96.0 & 88.0 & \textbf{\underline{83.5}} & 90.9 \\
    \bottomrule
  \end{tabularx}
  \vspace{-0.2cm}
\end{table}

\vspace{-0.2cm}
\paragraph{Robustness to Corruptions}
We further assessed the robustness of Inhibited Self-Attention to image corruptions using the ImageNet-C and ImageNet-$\bar{\text{C}}$ benchmarks.
We focus on the relative mean Corruption Error (rmCE), which normalizes corruption performance by each model's own clean error rate, thereby isolating robustness gains from differences in clean accuracy. ISA achieves the best rmCE across all model families and both benchmarks. Detailed results are provided in Appendix C.

\subsection{Results: Object Detection}  \label{sec:od}
We demonstrate the benefits of ISA beyond image classification on the COCO object detection benchmark~\cite{cocodataset}. Using ViT-DINO and DIFF-DINO as baselines, we compare their performance to ISA-DINO.

\paragraph{Attention-on-Detected-Objects (AoDO)}  
Unlike the ViT backbone, which employs dense attention, the official DINO implementation adopts Deformable Attention~\cite{xia2022vision} for improved efficiency, faster convergence, and more accurate localization~\cite{zhangdino,zhudeformable}. To analyze how inhibition affects the spatial focus of the decoder, we quantify the Attention-on-Detected-Objects (AoDO) for all 36,335 COCO objects. For each object, we consider the best predicted bounding box and its corresponding sampling locations, and compute AoDO as:
\vspace{-0.05cm}
\begin{equation}
    AoDO = \frac{\sum_{i \in \mathcal{F}} w_i}{\sum_{j=1}^{N} w_j},  
\quad \text{where} \quad
\mathcal{F} = \{i \mid \mathbf{p}_i \in \mathcal{B}_{\text{GT}}\}
\end{equation}

\noindent where $\mathbf{p}_i = (x_i, y_i)$ denotes the $i$-th sampling point, $w_i$ its attention weight, and $N=H \times L \times K$ the number of sampling points, with $H$ attention heads, $L$ feature pyramid levels, and $K$ points per head-level. $\mathcal{B}_{\text{GT}}$ is the ground-truth bounding box. AoDO extends the Energy Based Pointing game~\cite{wang2020score}, specifically for deformable attention. Given the availability of COCO segmentation maps, we additionally compute AoO using the corresponding masks, which more accurately capture attention focus on the actual object than the bounding boxes used in the Energy-Based Pointing Game and AoDO.

For objects detected by both the baseline and inhibited models ($\sim$87\% of all objects), we compare their AoO and AoDO values in ~\cref{tab:aoo}. ISA-DINO learns to sample more precisely within both the ground truth bounding box of the object of interest (AoDO), and the actual segmentation map (AoO). The Differential Attention mechanism in DIFF-DINO hurts both mAP and attention focus compared to the plain ViT backbone, which may be caused by its global subtraction distorting spatial relationships and undermining object localization. This highlights that strong ImageNet performance, potentially achieved via shortcuts such as texture or background bias, does not guarantee success on downstream tasks like object detection, which demands precise spatial localization and is inherently more resistant to such shortcuts.

\begin{table}[!t]
    
    \caption{Detection performance (mAP) and attention sampling focus on COCO and COCO-B with respect to segmentation maps (AoO) and ground truth boxes (AoDO).}
    \label{tab:aoo}
    \centering
    \scriptsize
    \setlength\tabcolsep{-1.5pt}

    \begin{tabular*}{0.95\columnwidth}{@{\extracolsep{\fill}} l c c c c c c}
    \toprule
    \multirow{2}{*}{Model} & \multicolumn{3}{c}{\textbf{COCO}} & \multicolumn{3}{c}{\textbf{COCO-B}} \\  \cmidrule{2-4}\cmidrule{5-7}
    & mAP (\%)  & AoO (\%)  & AoDO (\%) & mAP (\%) & AoO (\%)  & AoDO (\%)  \\
    \midrule
    DINO-S/16 & 38.2 & 42.5 & 62.3 & 37.5 & 42.7 & 62.4  \\
    DIFF-DINO-S/16 & 37.7 & 39.4 & 57.9 & 37.7 & 39.4 & 58.2 \\
    ISA-DINO-S/16 & \textbf{38.6} & \textbf{44.6}  &  \textbf{63.8} & \textbf{38.6} & \textbf{44.6}  &  \textbf{63.9}  \\ 

    \bottomrule

    \end{tabular*}
    \vspace{-0.3cm}
\end{table}

\paragraph{Background reliance on COCO-Background}

To assess the reliance of object detectors on background information, we introduce COCO-Background\footnote{The COCO-B dataset will be released on Zenodo upon publication.} (COCO-B), a variant of the COCO dataset where objects are placed against non-natural backgrounds sampled from the Places365 dataset. Conceptually similar to the Waterbirds dataset~\cite{sagawa2019distributionally}, COCO-B preserves object locations while replacing the original scene context, as explained below. This enables a direct evaluation of background and contextual dependence in object detection.

\begin{figure*}[!t]
\scriptsize
 \centering
    \begin{adjustbox}{width=0.95\linewidth}
    \begin{tikzpicture}[every node/.style={transform shape=false}]

        \node at (0,1.6) {\includegraphics[width=2.75cm]{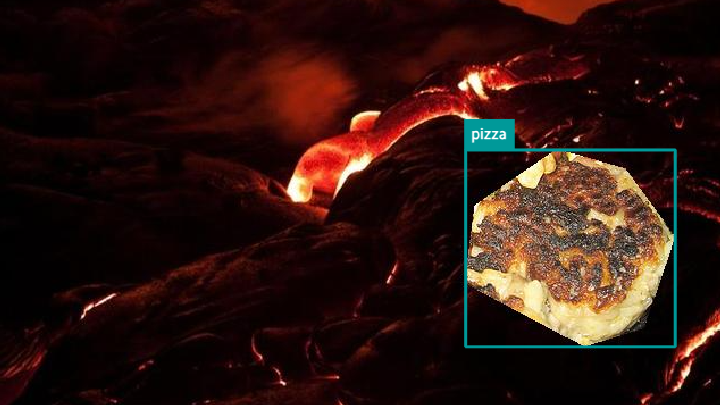}};
        \node at (2.8,1.6) {\includegraphics[width=2.75cm]{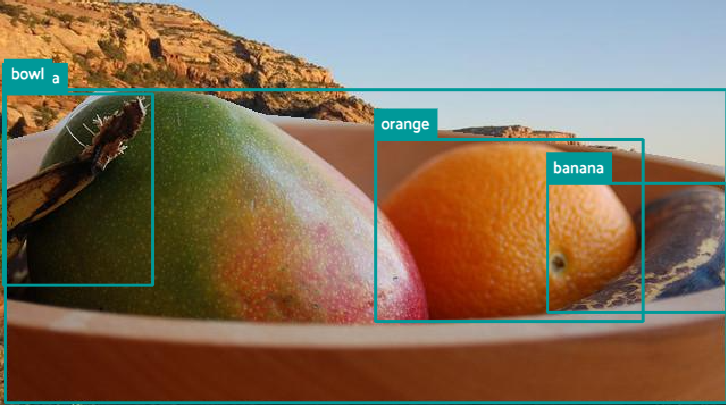}};
        \node at (5.6,1.6) {\includegraphics[width=2.75cm]{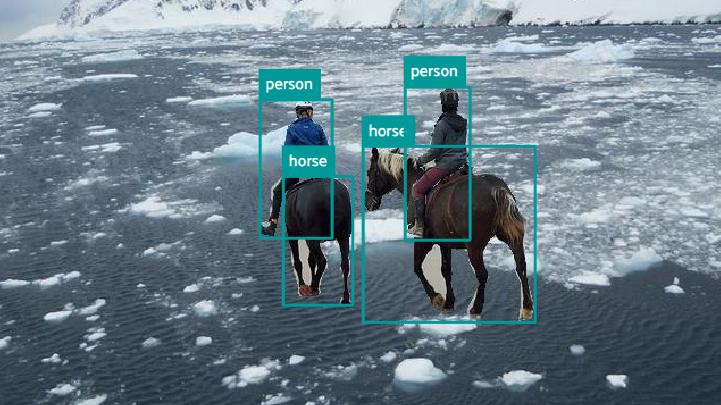}};
        \node at (8.4,1.6) {\includegraphics[width=2.75cm]{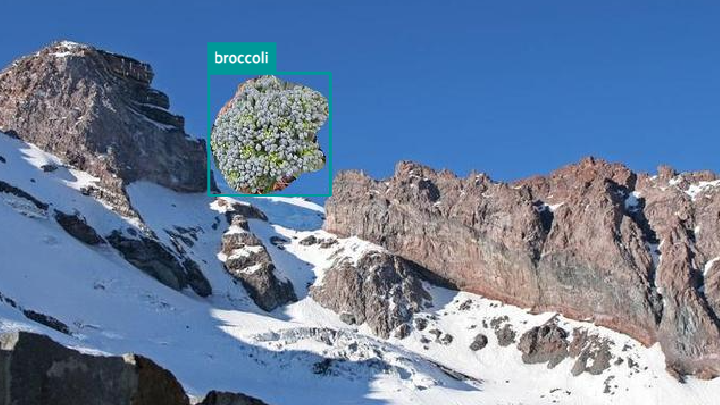}};
        \node at (11.2,1.6) {\includegraphics[width=2.75cm]{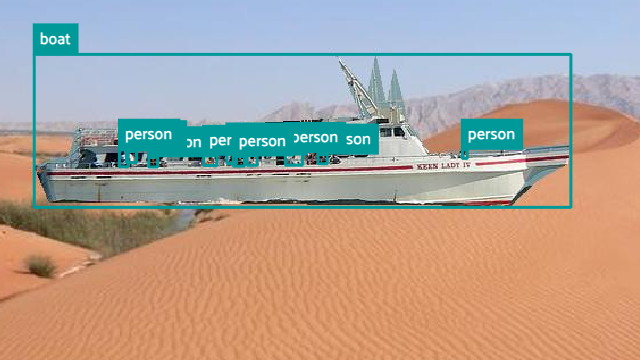}};
        \node at (0,0) {\includegraphics[width=2.75cm]{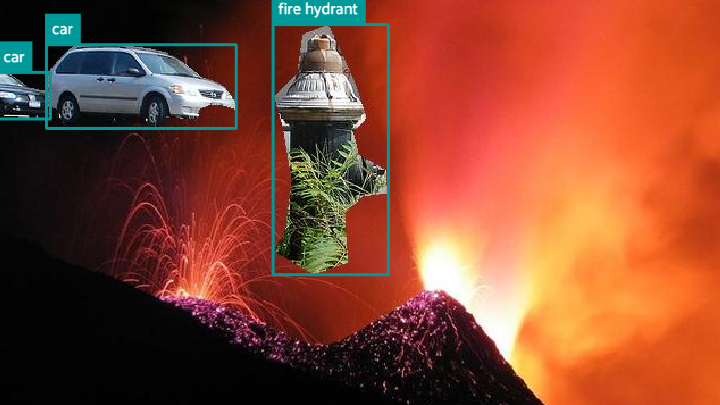}};
        \node at (2.8,0) {\includegraphics[width=2.75cm]{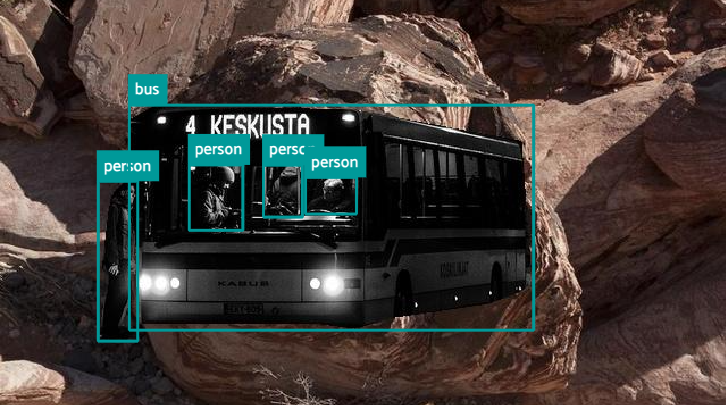}};
        \node at (5.6,0) {\includegraphics[width=2.75cm]{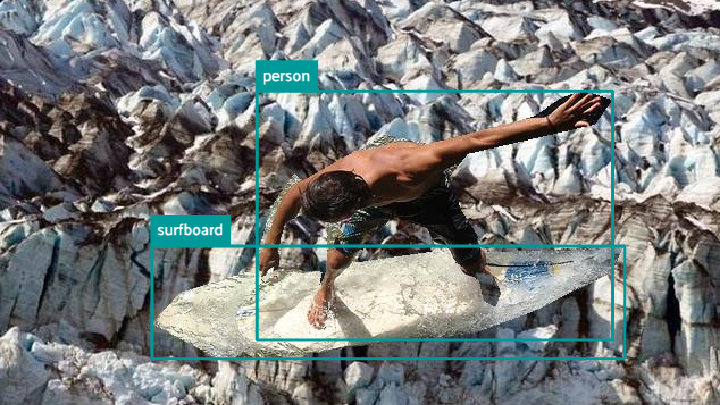}};
        \node at (8.4,0) {\includegraphics[width=2.75cm]{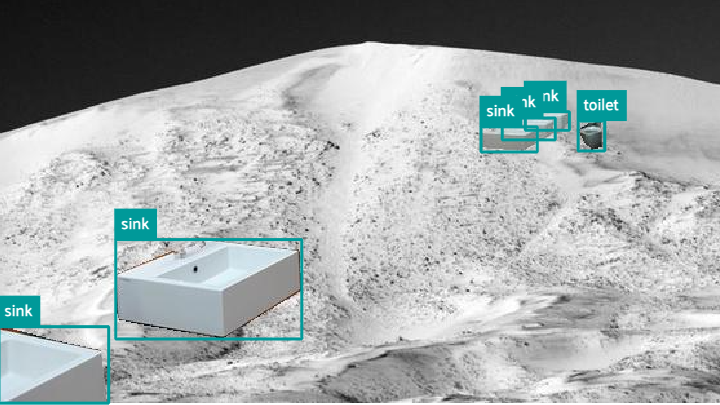}};
        \node at (11.2,0) {\includegraphics[width=2.75cm]{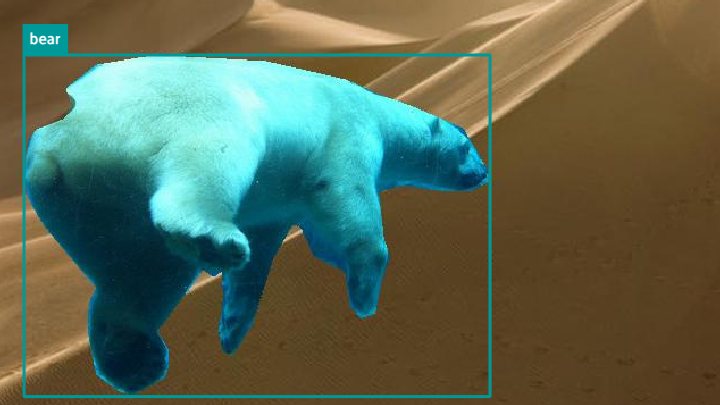}};

        \node at (0,   -1.1) {Volcano};
        \node at (2.8, -1.1) {Canyon};
        \node at (5.6, -1.1) {Glacier};
        \node at (8.4, -1.1) {Snowy Mountains};
        \node at (11.2,-1.1) {Desert Sand};

    \end{tikzpicture}
    \end{adjustbox}
    \caption{Example COCO-B images showing objects placed against non-natural backgrounds from each of the five classes: Volcano, Canyon, Glacier, Snowy Mountains, and Desert Sand. Ground truth bounding boxes and labels are visualized in blue.}
    \label{fig:cocob}
    \vspace{-0.3cm}
\end{figure*}
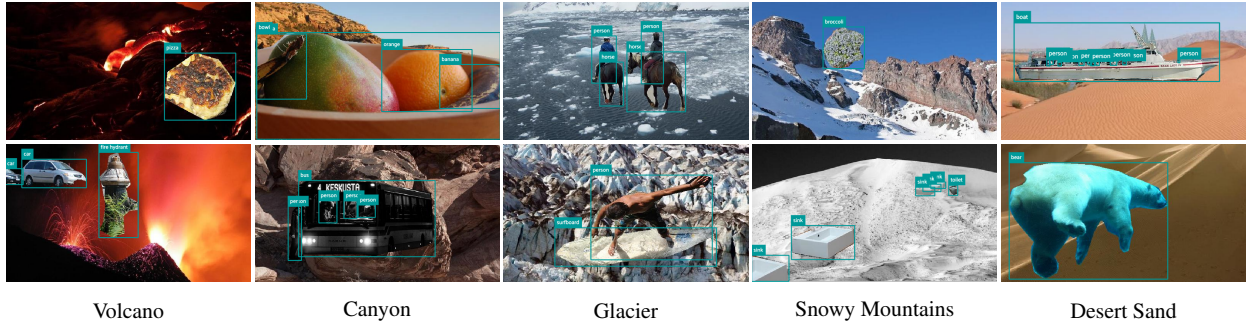

We selected five background classes, namely Volcano, Canyon, Glacier, Snowy Mountains, and Desert Sand, which are unlikely to produce false positives for the majority of COCO objects. Background images containing humans were manually removed to avoid introducing spurious detections. The dataset was generated by taking all non-empty validation images in COCO and compositing the objects onto randomly sampled backgrounds from each class, while keeping the use of each background class balanced. Backgrounds were resized and center-cropped to match the original image dimensions, ensuring full coverage without distortion or empty regions. Example images from COCO-B are shown in \cref{fig:cocob}.

In~\cref{tab:aoo}, we report results on background reliance. ISA-DINO maintains mAP under background alterations, whereas the model without inhibition experiences a substantial drop. DIFF-DINO shows no degradation but underperforms throughout, suggesting a lower performance ceiling rather than robustness. This indicates that the proposed inhibition effectively increases object-centric focus and reduces background reliance in object detection.

\subsection{Discussion}
With our proposed inhibited self-attention we address a key architectural limitation of standard attention mechanisms, namely the insufficient ability to suppress irrelevant features that usually lead to shortcut learning and biased predictions. We demonstrated that the focus of ViT attention can be sharpened by including inhibition. 
 The improved selectivity of ViT that we achieved, driven by the effective suppression of irrelevant cues in the computation of the self-attention of tokens, is a key factor for developing more reliable models. It indeed contributes to reducing the reliance on spurious shortcuts and background correlations, which prompt biased model predictions. By promoting selective attention to truly informative regions, inhibition enhances the model ability to make causally grounded and generalizable decisions rather than depending on coincidental contextual cues, as demonstrated by the consistently improved performance on ImageNet-R. Moreover, the resulting relevance maps exhibit clearer focus patterns, pointing towards more interpretable inference and allowing for a more transparent understanding of what drives the model predictions. 
 
 Our experiments focused on Vision Transformer models, while future work could extend this analysis to larger models to further investigate the effects of inhibition in higher-capacity architectures. 

\section{Conclusion}
We introduced Inhibited Self-Attention, a novel drop-in attention mechanism that enhances selectivity in Vision Transformers. We redesigned the self-attention mechanism by integrating an inhibitory process that sharpens the focus of self-attention to objects of interest, thus increasing the reliability of vision models. By leveraging negative attention scores, typically discarded by the \emph{softmax} function, our approach applies contextual inhibition to suppress irrelevant features, thus sharpening attention on objects of interest.

To quantify this improved object-centered focus, we introduced a new metric called \emph{Attention-on-Objects} (AoO), which measures how effectively models concentrate attention on target regions. ViTs with Inhibited Self-Attention produce attention maps with more precise and concentrated focus on class-relevant regions while reducing the impact of spurious cues. Our models assign significantly more attention to foreground objects than conventional ViTs, for both classification and detection tasks. The improved selectivity enhances decision-making by reducing reliance on shortcuts and spurious correlations. Furthermore, the improved focus on object-relevant regions translates to better generalization to out-of-distribution data, as reflected by higher robustness on ImageNet-R. Extending beyond classification benchmarks such as ImageNet and Waterbirds, ISA models also achieved improved results on COCO and COCO-B, showing more accurate object detection, better focus on target objects, and reduced reliance on backgrounds. 

\bibliographystyle{elsarticle-num}
\bibliography{main.bib}
\end{document}